\documentclass[journal]{IEEEtran}
\usepackage{amsmath,amsfonts}
\usepackage{algorithm}
\usepackage{algorithmic}
\usepackage{array}
\usepackage[caption=false,font=normalsize,labelfont=sf,textfont=sf]{subfig}
\usepackage{textcomp}
\usepackage{stfloats}
\usepackage{url}
\usepackage{verbatim}
\usepackage{graphicx}
\usepackage{cite}
\usepackage{makecell}
\usepackage{color}

\begin{document}

\title{Convergence-Latency-Aware Adaptive Modulation and Resource Allocation in RIS-Assisted Wireless Federated Learning}
%Ris-assisted Adaptive Modulation and Resource Allocation in Federated Learning 
\author{Liwei Wang, Wen Chen, Jun Li, Qingqing Wu, Ming Ding, Xusheng Zhu and Qiong Wu
\thanks{Liwei Wang, Wen Chen and Qingqing Wu are with the Broadband Access
Network Laboratory, Shanghai Jiao Tong University, Minhang 200240, China
(e-mail: wanglw2000; wenchen; qingqingwu@sjtu.edu.cn). 

Jun Li is with the School of Information Science and Engineering, Southeast University, Nanjing 210096, China (e-mail: jun.li@seu.edu.cn).

Ming Ding is with Data61, CSIRO, Sydney, NSW 2015, Australia (e-mail:
ming.ding@data61.csiro.au).

Xusheng Zhu is with the Department of Electronic
and Electrical Engineering, University College London, WC1E 6BT London,
U.K. (e-mail: xusheng.zhu@ucl.ac.uk).

Qiong Wu is with the School of Internet of Things Engineering, Jiangnan
University, Wuxi 214122, China (e-mail: qiongwu@jiangnan.edu.cn).}

}

% The paper headers
\markboth{IEEE TRANSACTIONS ON COGNITIVE COMMUNICATIONS AND NETWORKING}%
{Shell \MakeLowercase{\textit{et al.}}: A Sample Article Using IEEEtran.cls for IEEE Journals}

% \IEEEpubid{0000--0000/00\$00.00~\copyright~2021 IEEE}
% Remember, if you use this you must call \IEEEpubidadjcol in the second
% column for its text to clear the IEEEpubid mark.

\maketitle

\begin{abstract}
Federated learning (FL) over wireless networks suffers from significant training latency and degraded convergence due to unreliable wireless transmission, especially under blocked propagation environments. Although reconfigurable intelligent surfaces (RISs) can improve communication reliability, existing wireless FL studies rarely characterize the trade-off between learning convergence and communication delay under modulation-dependent transmission errors. In this paper, we consider a wireless FL system operating under RIS-assisted blocked-link propagation scenarios, and focus on adaptive modulation and sub-channel allocation for convergence-latency aware communication design. By characterizing the effect of symbol errors on uploaded local gradients, we derive a convergence-related upper bound that reveals the impact of symbol error rate (SER) on FL loss decay. Based on this result, we formulate a joint convergence-latency optimization problem, which is cast as a mixed-integer nonlinear programming (MINLP) problem, and solve it using a low-complexity hybrid alternating optimization framework. Extensive experiments on MNIST, CIFAR-10, and Speech Commands show that the proposed scheme consistently achieves faster convergence and higher test accuracy than existing adaptive communication schemes, especially in complex tasks and challenging wireless scenarios.
\end{abstract}


\section{Introduction}
\IEEEPARstart{W}{ith} the proliferation of artificial intelligence (AI) applications in domains such as healthcare, finance, and autonomous systems, distributed machine learning frameworks have gained increasing traction\cite{jing2022}. Traditional machine learning approaches involve uploading all raw data to a central server, where centralized model training is then performed on the aggregated dataset. Nevertheless, the traditional centralized learning framework faces the challenge of data isolation, due to the increasing emphasis on privacy issues\cite{yang2019federated}.  Federated learning (FL), in particular, has emerged as a privacy-preserving learning paradigm, enabling collaborative model training across distributed devices without exposing raw data\cite{mcmahan2017}. FL typically necessitates a substantial number of communication iterations to achieve the target model accuracy, a phenomenon that becomes particularly pronounced when the training process involves a relatively large cohort of participating clients \cite{Nguyen2021}. Consequently, the model performance and the latency induced by unreliable wireless transmissions and heterogeneous local computation capacities across clients emerge as a critical bottleneck in wireless FL systems \cite{deng2023}. Moreover, although computational efficiency may be optimized, the inherent long-range and blocked communication between edge devices and remote cloud servers inevitably incurs significant transmission delays and errors, thereby compromising client quality-of-service (QoS).

Latency is one of the most critical performance bottlenecks in wireless FL systems. The local computation delay depends on the data size and computational capability of participating devices, while the uplink transmission delay is influenced by channel conditions, modulation schemes, and allocated bandwidth. The downlink broadcasting delay, on the other hand, is typically smaller compared with the previous two items. Existing studies have investigated several strategies to reduce latency in wireless FL\cite{vtc2024,kang2022,chen2024,wang2024,han2025,Tan2025, cheniot2024 }. The authors in \cite{vtc2024} jointly optimize the computation frequency and communication bandwidth while respecting the energy budget to minimize the total FL execution time. In \cite{kang2022}, the authors address the latency-minimization problem in FL, explicitly accounting for  potentially different privacy protections and data imbalance. Also, model quantization and compression techniques have been proposed to reduce the size of transmitted updates\cite{chen2024}, thereby shortening transmission time. A novel FL aggregation mechanism is proposed in \cite{wang2024} to reduce the communication overhead, by selecting more valuable layers to upload by their feedback framework. The authors in \cite{han2025}, \cite{Tan2025} propose a dynamic quantization framework that jointly minimizes per-client energy expenditure and improves uplink communication efficiency by adaptively tuning the quantization levels. \cite{cheniot2024} dynamically assigns heterogeneous quantization resolutions across clients and correspondingly re-weighted their scheduling probabilities, in order to match the resulting uplink heterogeneity and latency optimization. The above works broaden the scope of applying latency optimization in FL. However, these works often focus too much on communication delay and neglect the model accuracy problems caused thereby.

Unreliable communication is also a major challenge that FL faces in practical deployment. The core of FL is distributed collaboration, however, unreliable communication will directly affect its learning efficiency and speed, as well as the quality and convergence of the model\cite{jsac2021}. A lot of works have been done to alleviate the negative impacts caused by unreliable wireless communication in FL \cite{Mingzhechen2021, jing2025,globecom2024,zhao2023, sun2025 }. The authors in \cite{Mingzhechen2021}, \cite{jing2025} propose a novel FL framework in which the impact of the wireless packet transmission errors is considered to select clients for a better model performance. A FL framework, FL-GC, is presented in \cite{globecom2024}, where the server infers the local gradient information of clients with transmission failures or those not selected for participation, utilizing locally uploaded gradients from prior communication rounds. Also, recently reconfigurable intelligence surface (RIS) is regarded as one of the key technologies to address the unreliable issue in future wireless communication\cite{Basar2019}, \cite{zxs2024}. A large amount of work has already applied the RIS to address the research issues related to unreliable communication in FL. The authors in \cite{zhao2023} propose a novel performance-centric long-term design scheme that integrates multiple communication rounds, with the primary objective of minimizing the optimality gap associated with the loss function in AirComp via RISs. By converting SINR thresholds to SER to capture key performance constraints, \cite{sun2025} selects local models within an acceptable error range for global aggregation, enabling FL to incorporate more tolerable-error local models and gain performance improvements. Existing efforts have been devoted to achieve higher model accuracy at the cost of increased power consumption and latency, in an attempt to mitigate the impact of unreliable communication as much as possible.

Due to the limited resources of wireless FL, some works focus on achieving better model performance and higher learning efficiency through client scheduling. The authors in \cite{zeyu2025} and \cite{han2024} formulate a joint optimization problem of resource allocation and client scheduling, to minimize the energy consumption in the social network scenario and the BS-client scenario respectively.
Also,  a joint bandwidth allocation and scheduling framework is built in \cite{icc2020} to achieve a good trade-off between the learning efficiency and latency. The work in \cite{fan2023} develops a client scheduling and resource allocation method to minimize the training delay in a novel client mobility network. Moreover, the client scheduling framework is delineated in \cite{ren2020} to achieve the optimal trade-off between channel quality and local data importance.

To sum up, existing studies have investigated latency reduction, communication reliability, and resource management in wireless federated learning. However, the trade-off between convergence and latency under modulation-dependent transmission errors is still not well understood. Motivated by this issue, we consider a wireless FL system where RISs are deployed to assist blocked NLoS transmissions, and focus on adaptive modulation and sub-channel allocation under modulation-dependent transmission errors. We characterize the impact of symbol errors on FL convergence, formulate a joint convergence--latency optimization problem, and develop a corresponding solution framework.

The main contributions of this paper are summarized as follows.

\begin{itemize}
    \item We consider a wireless FL system under direct LoS and RIS-assisted blocked NLoS transmission scenarios, and establish a communication model that jointly characterizes uplink transmission latency and gradient distortion caused by symbol errors. The proposed model explicitly reveals the reliability-latency trade-off caused by adaptive modulation in wireless FL.

    \item We derive a convergence-related upper bound that links modulation-dependent symbol errors with the expected FL loss decay. The analysis quantifies how the SER affects gradient aggregation errors and further degrades the convergence behavior, thereby providing an analytical basis for convergence-latency aware modulation and resource allocation.

    \item We formulate the joint adaptive modulation and subchannel allocation problem as a mixed-integer nonlinear programming (MINLP) problem and propose a low-complexity hybrid alternating-optimization framework to solve it efficiently. Specifically, continuous relaxation and Newton iterations are adopted for modulation optimization, while binary relaxation and KKT-based optimization are employed for sub-channel allocation. Experiments on MNIST, CIFAR-10, and Speech Commands demonstrate that the proposed design achieves faster convergence and higher accuracy than benchmark schemes.
\end{itemize}

% The rest of this paper is organized as follows. Section II presents the system model, including the FL model, transmission model, channel error model, and adaptive modulation mechanism. Section III analyzes the convergence behavior of FL under transmission errors and formulates the joint optimization problem. Section IV develops the corresponding optimization methods for modulation order design and sub-channel allocation. Section V provides experimental results and performance comparisons under different datasets and wireless settings. Finally, Section VI concludes this paper.
\section{System Model}
In this section, we present the considered wireless FL system, where RISs are deployed only to assist blocked NLoS links. We then introduce the learning model, wireless transmission model, channel error model, and adaptive modulation mechanism.

\subsection{FL Model}
\begin{figure*}[!t]
\centering
\includegraphics[width=0.70\textwidth]{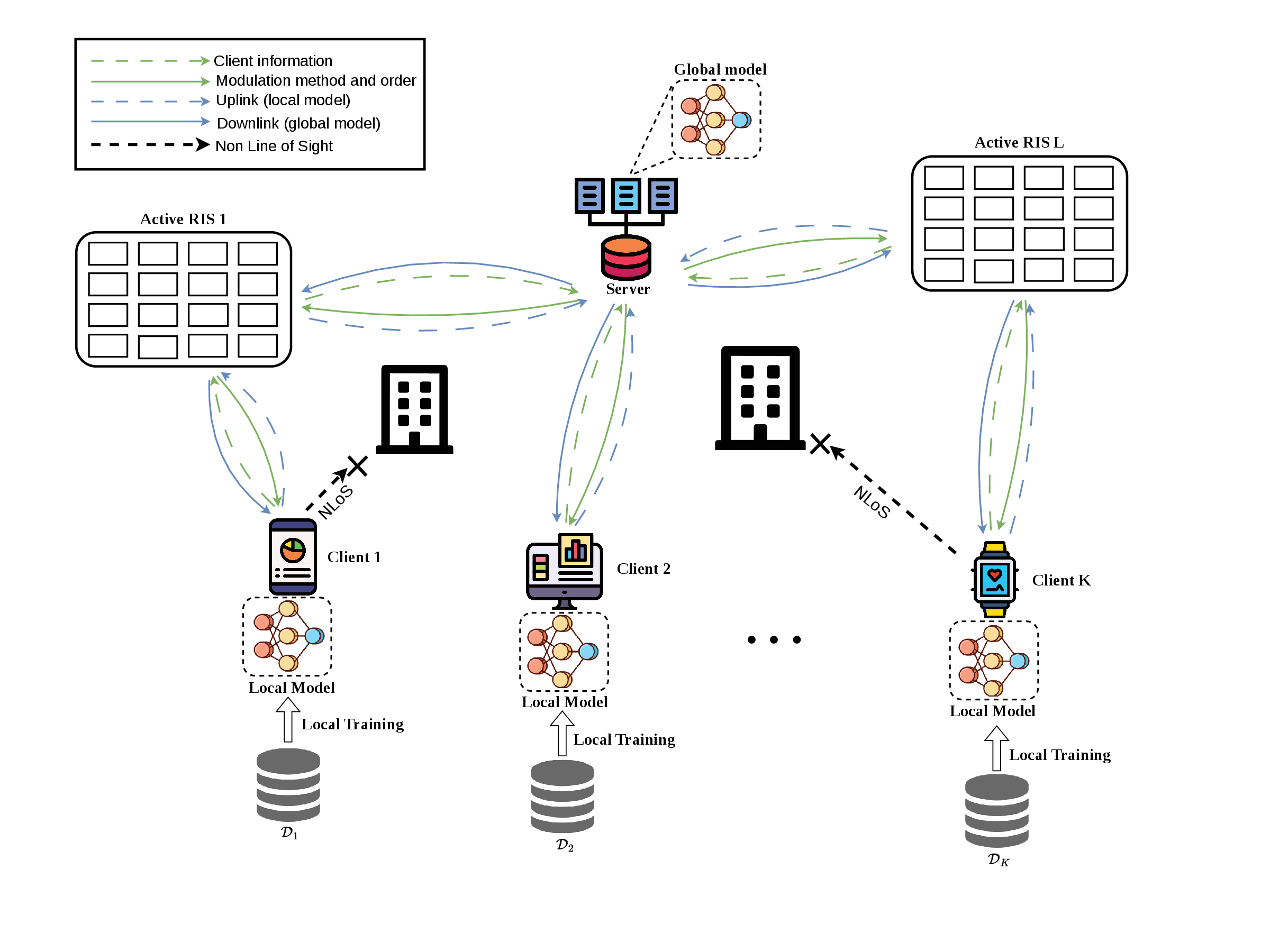}%
\caption{System model of the RIS-assisted wireless FL system.}
\label{sysmodel}
\end{figure*}

As shown in Fig. \ref{sysmodel}, we consider an RIS-assisted wireless FL system consisting of one parameter server (PS), $K$ edge clients, and $L$ RISs with $R$ reflecting elements. For clients whose direct links to the PS are blocked, the uplink transmission is assisted by RIS reflection; otherwise, the direct LoS link is used. Each client $k\in\mathcal{K}=\{1,2,\ldots,K\}$ owns a local dataset $\mathcal{D}_k$ with size $D_k=|\mathcal{D}_k|$, where each sample is denoted by $(u_i,v_i)$.
% As shown in Fig. \ref{sysmodel}, the RIS-assisted FL system under consideration comprise: 1) one parameter server (PS), 2) $K$ edge clients who cooperatively perform an FL algorithm, and 3) L RISs with $R$ active reflecting elements are deployed to enhance blocked NLoS links between clients and the PS. In our scenario, for clients whose direct links to the PS are blocked, the uplink signal is assumed to be assisted by RIS reflection; otherwise, the direct LoS link is used. Consider edge client $k \in \mathcal{K} = \{1,2,\cdots,K\}$ has its own i.i.d. sampled local dataset $\mathcal{D}_k$ with data $\{(\boldsymbol{u}_i,v_i)\}_{i=1}^{\lvert \mathcal{D}_k \rvert} \in \mathcal{D}_k$, where $\boldsymbol{u}_i$ represents the $i$-th training data sample and $v_i$ represents the corresponding ground-truth label. To train data while protecting the clients' privacy, local clients transmit the gradients or parameters instead of raw local data via wireless channels. The global gradients or parameters are then obtained through average aggregation at the PS. 

For a given model parameter $\boldsymbol{\theta}$, the local loss function for client $k$ is given by
\begin{equation}
    F_k(\boldsymbol{\theta}) = \frac{1}{D_k} \sum_{(\boldsymbol{u}_i,v_i)\in \mathcal{D}_k} f(\boldsymbol{\theta};\boldsymbol{u}_i,v_i),
\end{equation}
where $D_k=\lvert \mathcal{D}_k \rvert$ that represents the dataset size of the $k$-th client. $f(\boldsymbol{\theta};\boldsymbol{u}_i,v_i)$ is the loss function, which is denoted by $f_k(\boldsymbol{\theta})$ hereafter for brevity. The loss function captures different performance of FL algorithm when applying different learning task. We assume that clients have unequal dataset size in the local training, and the global loss function with model parameter $\boldsymbol{\theta}$ is defined as
\begin{equation}
    F(\boldsymbol{\theta}) = \frac{1}{D_{all}}\sum_{k=1}^K D_k F_k(\boldsymbol{\theta}),
\end{equation}
where $D_{all} = \sum_k D_k$. The global loss function is weighted average of the local loss function by the dataset size of each client. The FL process aims to optimize the model parameter $\boldsymbol{\theta}$ that minimizes $F(\boldsymbol{\theta})$, i.e.,
\begin{equation}
    \boldsymbol{\theta}^* = \mathop{\arg\min}\limits_{\theta} F(\boldsymbol{\theta}).
\end{equation}

To facilitate cooperative training among edge devices, each client independently computes its local gradients by minimizing $F_k(\boldsymbol{\theta})$ in parallel, and subsequently, the server aggregates these local gradients to derive the global gradient In the $t$-th communication round, the $k$-th client evaluates the gradient of its local model with its corresponding local dataset $\mathcal{D}_k$, as expressed by
\begin{equation}
    \boldsymbol{g}_k^t = \nabla F_k(\boldsymbol{\theta}^t) =\frac{1}{D_{k}} \sum_{(\boldsymbol{u}_i,v_i)\in \mathcal{D}_k}  \nabla f_k(\boldsymbol{\theta}^t) .
\end{equation}

After receiving the updates of all clients, the PS aggregates all local gradients and computes the global gradient, which is updated as 
\begin{equation}
    \boldsymbol{g}^t = \nabla F(\boldsymbol{\theta}^t)=\frac{1}{D_{all}}\sum_{k=1}^K D_k \boldsymbol{g}_k^t.
    \label{eq:aggregation}
\end{equation}

Then based on the global gradient $\boldsymbol{g}^t$, the PS executes a global model update utilizing the gradient descent algorithm  in order to derive an updated global model, which is given by
\begin{align}
    \boldsymbol{\theta}^{t+1} &= \boldsymbol{\theta}^{t}-\eta \boldsymbol{g}^{t},
    \label{eq:gloal update}
\end{align}
where $\eta$ denotes the learning rate. 

Due to wireless channel impairments, the uploaded gradients may be distorted by transmission errors. The impact of such errors on FL convergence will be analyzed in Section III.
% Following this, the PS broadcasts the updated global model $\boldsymbol{\theta}^{t+1}$ to all participating clients, initiating the subsequent round of iterative training. This process continues until the model converges to an optimal solution. Local models are transmitted through a wireless channel, and in practical scenarios, the PS may receive these local models with erroneous symbols due to channel fading and noise inherent in wireless communication. We will present our model for the communication scenario in the following subsection and clarify the effect of transmission errors on FL performance.

% \begin{table}[!t]
% \caption{LIST OF NOTATION\label{tab:table1}}
% \centering
% \renewcommand\arraystretch{1.25}
% \begin{tabular}{|c||c|}
% \hline
% \makebox[0.1\textwidth][c]{\textbf{Notation}} & \makebox[0.3\textwidth][c]{\textbf{Description}}\\
% \hline
% $K$ & Number of clients \\
% \hline
% $L$ & Number of RISs \\
% \hline
% $R$ & Element number of a RIS \\
% \hline
% $(\boldsymbol{u}_i,v_i)$ & Data sample and label collected by clients \\
% \hline
% $\mathcal{K}$ & Set of clients \\
% \hline
% $\mathcal{D}_k$ &  Dataset held by the client $k$ \\
% \hline
% $D_k$ &  Dataset size of the client $k$ \\
% \hline
% $D_{all}$ &  Dataset size of all clients \\
% \hline
% $\boldsymbol{\theta}^t$ &  \makecell[c]{Global model \\at the $t$-th communication round} \\
% \hline
% $\boldsymbol{\theta}_k^t$ &  \makecell[c]{Local model of client $k$ \\at the $t$-th communication round} \\
% \hline
% $a_{k,n}$ &  Sub-channel allocation indicator \\
% \hline
% $M_{k,n}$ &  Modulation order \\
% \hline
% $Z$ &   Total element number of learning parameters \\
% \hline
% $\epsilon$ &   \makecell[c]{Average error of\\ an element with one symbol error} \\
% \hline
% $\Delta$ &   \makecell[c]{Expectation of the average\\ error for the $k$-th client and the $z$-th
% element}\\
% \hline
% $\sigma$ &   \makecell[c]{Number of symbols contained in an element}\\
% \hline

% \end{tabular}
% \end{table}

\subsection{Transmission Model}
In our system, the global and local gradients are transmitted via the wireless channel with or without RIS assistance, and we do not consider the reflection time of the signal on RIS. It is considered that orthogonal frequency-division multiple access (OFDMA) is employed for accessing the uplink channel. In the case, the system bandwidth $B$ is divided into $N$ sub-channels, which are assigned to clients without interference. Define $B_n$ as the bandwidth of the sub-channel $n$ and we have $\sum_{n=1}^N B_n \leq B$. Since all clients are required to upload their local gradients in each communication round, each client should be allocated at least one sub-channel. Accordingly, we assume that the number of sub-channels satisfies $N\ge K$. Given that the downlink bandwidth is accessible to all clients and the entire system bandwidth can be applied to downlink transmission, we limit our analysis to channel errors occurring only during uplink transmission. We assume that the channel state information is able to be known by the channel estimation algorithm in each communication round.
\subsubsection{Uplink Transmission}
Considering the modulation order, the uplink data rate of client $k$ can be expressed as
\begin{equation}
    r_k^{U}(\boldsymbol{a}_k, \boldsymbol{M}_k) = \sum_{n=1}^N a_{k,n} B_n \log_2 (M_{k,n}),
\end{equation}
where $\boldsymbol{a}_k=[a_{k,1},a_{k,2},\cdots,a_{k,N}]$ is a sub-channel allocation vector, $a_{k,n}\in \{0,1\}$ and $a_{k,n}=1$ indicates that sub-channel $n$ is allocated to client $k$, and $a_{k,n}=0$, otherwise. $\boldsymbol{M}_k$ is the modulation order vector and $M_{k,n}$ is the modulation order of the client $k$ to transmit data on sub-channel $n$.

Then let $Z$ denote the total number of learning parameters and $\phi$ denote the quantitative bit number for each parameter. Note that the amount of feedback client information data is small and the communication delay is negligible. Subsequently, the data size of the local gradient can be assessed as $\phi \times Z$, and the latency associated with uploading the local gradient is defined by
\begin{align}
    T_k^{U} &= \frac{\phi Z}{r_k^{U}(\boldsymbol{a}_k, \boldsymbol{M}_k)} \notag\\
    &= \frac{\phi Z}{\sum_{n=1}^N a_{k,n} B_n \log_2 (M_{k,n})}.
\end{align}
\subsubsection{Downlink Transmission}
For the downlink channel, we assume that all downlink bandwidth is occupied to broadcast the global model. With BPSK modulation, the achievable downlink data rate can be expressed as
\begin{equation}
    r^D = B.
\end{equation}

Meanwhile, due to the fact that each gradient has its corresponding model parameter, the total number of parameters in the global model to broadcast is $Z$ as well. Then the downlink latency for the global model broadcasting is given by, with the $\phi$-bits quantization,
\begin{align}
    T^{D} = \frac{\phi Z}{B}.
\end{align}
\subsubsection{One Round Latency}
In addition to the uplink and downlink transmission delays, we also consider the latency of local model updates. Let $f_k$(cycle/s) denote the computation capability of client $k$. Moreover, we define $C$ as the total number of floating-point operations involved in backpropagation with one individual data sample. Then, with $D_k$, the local gradient calculation latency of the client $k$ is
\begin{equation}
    T_k^C = \frac{D_k C}{f_k}.
\end{equation}

Due to the high computation power of the PS and the small amount of computation of the gradient weighted aggregation, we ignore the time for global model update. Moreover, the PS will not perform the aggregation process until all gradients from the clients are received. Accordingly, the one round latency can be expressed as
\begin{equation}
    T = \max_k\{T_k^C + T_k^U\}+T^D.
\end{equation}
\subsection{Channel Error Model}
In our paper, for both direct LoS and RIS-assisted blocked NLoS links, the SER depends on the adopted modulation scheme. Given that the downlink bandwidth is accessible to all clients and allows for the use of more robust channel coding, we focus solely on channel errors occurring during uplink transmission and use the nominal symbol-rate-limited throughput. There are two common modulation schemes that will be considered in this paper, MPSK and MQAM. We denote the signal-to-noise ratio (SNR) of client $k$ on sub-channel $n$ is $\gamma_{k,n}=\frac{p_{k,n}|h_{k,n}|^2}{N_0B_n}$, where $p_{k,n}$ is the transmit power of client $k$ on sub-channel $n$, $h_{k,n}$ is the corresponding channel coefficient, and $N_0B_n$ denotes the noise power over sub-channel $n$. Then, we characterize the SER under two propagation conditions: direct LoS transmission and RIS-assisted blocked NLoS transmission.  
\subsubsection{LoS scenarios}
In this case, we consider that there are no obstructions between the straight channel path from this client to the PS. Following \cite{proakis2008digital}, when MPSK is applied, we express the SER of client $k$ on sub-channel $n$ as
% \begin{equation}
%     q_{k,n,M}^{LoS,PSK}\leq2 Q\left(\sqrt{\frac{2 h_{k,n}^2 E_{k,n}\log_2 (M_{k,n})}{N_0}}\sin \frac{\pi}{M_{k,n}}\right),
% \end{equation}
\begin{equation}
q_{k,n,M}^{\mathrm{LoS,PSK}}
\le
2Q\left(
\sqrt{
2\gamma_{k,n}\log_2(M_{k,n})
}
\sin\left(\frac{\pi}{M_{k,n}}\right)
\right),
\end{equation}
where $0\leq q_{k,n,M}^{LoS,PSK} \leq 1$, $\forall k \in K$, $h_{k,n}$ is the
channel gain between the client $k$ and the PS on sub-channel $n$, and $Q$-function $Q(x)= \frac{1}{\sqrt{2\pi}}\int_x^\infty \exp \left(-\frac{u^2}{2}\right)du$ that denotes the tail function of the standard normal distribution. 

Moreover, when MQAM is applied, we express the SER of client $k$ on sub-channel $n$ as
% \begin{equation}
%     q_{k,n,M}^{LoS,QAM}\leq4Q\left(\sqrt{\frac{3 h_{k,n}^2 E_{k,n} \log_2^2 (M_{k,n}) }{\left(M-1\right)N_0}}\right),
% \end{equation}
\begin{equation}
q_{k,n,M}^{\mathrm{LoS,QAM}}
\le
4Q\left(
\sqrt{
\frac{3\gamma_{k,n}\log_2(M_{k,n})}
{M_{k,n}-1}
}
\right),
\end{equation}
where $0\leq q_{k,n,M}^{LoS,QAM} \leq 1$, $\forall k \in K$. Hence, we can express the SER of client $k$ in LoS scenarios as 
\begin{equation}
    q_{k}^{LoS,PSK(QAM)} = \frac{\sum_{n=1}^{N} a_{k,n} q_{k,n,M}^{LoS,PSK(QAM)}}{\sum_{n=1}^{N} a_{k,n}}.
\end{equation}
\subsubsection{NLoS with RIS-assisted scenarios}
In this scenario, we assume there are buildings blocking the channel between the client and the PS, which greatly weakens the transmission strength of the signal. We assume all NLoS clients transmit their signal via RISs, as MPSK is selected, then the SER of client $k$ on sub-channel $n$ is \cite{Basar2019}
\begin{equation}
    q_{k,n,M}^{NLoS,PSK} = \frac{1}{\pi} \int_0^{\frac{(M-1)\pi}{M}} \mathbb{M}_\gamma\left(\frac{-\sin^2\left(\frac{\pi}{M}\right)}{\sin^2 \alpha}\right) d\alpha,
    \label{eq:qnlos_psk}
\end{equation}
where $\mathbb{M}_\gamma$ is the moment generating function (MGF) as \cite{proakis2008digital}
\begin{equation}
    \mathbb{M}_{\gamma }(s)= \left ({\dfrac {1}{1-\frac {sL(16-\pi ^2) \gamma_{k,n}}{8}}}\right) ^{\frac {1}{2}} \! \!\exp \left ({\dfrac { \frac {sL^{2} \pi ^{2} \gamma_{k,n}}{16}}{1-\frac {sL(16-\pi ^{2}) \gamma_{k,n}}{8}} }\right).
\end{equation}

Since $\mathbb{M}_\gamma$ is an increasing function of $s$, \eqref{eq:qnlos_psk} can be upper bounded by setting $\alpha = \frac{M-1}{M}\pi$, which yields
\begin{align}
     &q_{k,n,M}^{NLoS,PSK} \notag\\
     \leq& \frac{M-1}{M} \left(\dfrac{1}{1+\frac{\beta L (16-\pi^2)\gamma}{8}}\right)^{\frac{1}{2}} \exp \left(\dfrac{\frac{-\beta L^2 \pi^2 \gamma}{16}}{1 + \frac{\beta L (16-\pi^2)\gamma}{8}} \right),
     \label{eq:qnlos_pskleq}
\end{align}
where $\beta = \frac{\sin^2\left(\pi/M\right)}{\sin^2\left((M-1)\pi/M\right)}$. From \eqref{eq:qnlos_pskleq}, we can see that with RIS-assisted, even though the SNR $\gamma$ is relatively low, the average SER remains rather low.

Moreover, by utilizing the above-mentioned MGF, we are also able to derive the SER for square of client $k$ on sub-channel $n$ for MQAM constellations \cite{Basar2019}
\begin{small}
\begin{align}
    &q_{k,n,M}^{NLoS,QAM} \notag\\
    =&\frac {4}{\pi } \left ({1-\frac {1}{\sqrt {M}} }\right) \int _{0}^{\pi /2} \mathbb{M}_{\gamma } \left ({\frac {-3}{2(M-1) \sin ^{2} \! \alpha }}\right) d\alpha \notag\\
    &\qquad -\, \frac {4}{\pi } \left ({\!1-\frac {1}{\sqrt {M}} \!}\right)^{2} \int _{0}^{\pi /4} \mathbb{M}_{\gamma } \left ({\! \frac {-3}{2(M-1) \sin ^{2} \! \alpha }\!}\right) d \alpha.
    \label{eq:qnlos_qam}
\end{align}
\end{small}

Similarly, we can obtain an upper bound for \eqref{eq:qnlos_qam} by setting $\alpha=\frac{\pi}{2}$ in the first term and $\alpha=\frac{\pi}{4}$ in the second term respectively, which yields
\begin{small}
\begin{align}
    q_{k,n,M}^{NLoS,QAM} 
    \leq 2 &\left ({1-\frac {1}{\sqrt {M}} }\right) \mathbb{M}_{\gamma } \left ({\frac {-3}{2(M-1) }}\right) \notag \\
    &\qquad- \left ({\!1-\frac {1}{\sqrt {M}} \!}\right)^{2}  \mathbb{M}_{\gamma } \left ({\! \frac {-6}{M-1}}\right).
\end{align}
\end{small}

Then we can also express the SER of client $k$ in NLoS scenarios as 
\begin{equation}
    q_{k}^{NLoS,PSK(QAM)} = \frac{\sum_{n=1}^{N} a_{k,n} q_{k,n,M}^{NLoS,PSK(QAM)}}{\sum_{n=1}^{N} a_{k,n}}.
\end{equation}

% \subsection{Modulation Scheme Allocating Mechanism}
% In wireless FL, local gradients are transmitted over unreliable wireless channels, and symbol errors may distort the uploaded model updates, thereby degrading the aggregation accuracy and learning performance. A higher modulation order can improve the transmission rate and reduce communication latency, but it usually leads to a larger SER under the same channel condition. Conversely, a lower modulation order provides higher transmission reliability at the cost of reduced spectral efficiency. Therefore, an adaptive modulation mechanism is needed to balance communication latency and gradient reliability. In the following section, we analyze the impact of modulation-dependent symbol errors on FL convergence and formulate the corresponding convergence-latency optimization problem.

\section{Performance Analysis and Problem Formulation}
The modulation order affects both the uplink transmission latency and the SER-induced gradient distortion, which motivates the convergence-latency analysis.

In this section, we will first analyze the convergence behavior and how the channel error affects the performance of FL. To obtain a tractable convergence-related characterization of the impact of wireless transmission errors, we make several assumptions on the loss function $F(\boldsymbol{\theta})$. Then, the generalized optimality gap, which is suitable for generic wireless networks, is derived to characterize the learning efficiency between two arbitrary communication rounds. And the obtained optimality gap sheds light on how the imperfect gradient updates affect the convergence rate of FL. Next, we focus on the effect of RIS-assisted system channel error rate on FL convergence performance and formulate the corresponding optimization problem with adaptive modulation and sub-channel allocation mechanism.
\subsection{Performance Analysis}
For analysis, we make the following assumptions.

\textit{Assumption 1($\ell$-smooth):} The global loss function $F(\boldsymbol{\theta})$ is smooth, at any point $\boldsymbol{\theta}\in \mathbb{R}^d$, with positive constant $\ell > 0$, i.e.,
\begin{equation}
    \left\Vert \nabla F(\boldsymbol{\theta}) - \nabla F(\boldsymbol{\theta}^{\prime})  \right\Vert \leq \ell \left\Vert \boldsymbol{\theta}-\boldsymbol{\theta}^{\prime} \right\Vert, \quad \forall\boldsymbol{\theta},\boldsymbol{\theta}^{\prime} \in \mathbb{R}^d,
\end{equation}
where $\Vert\cdot\Vert$ is the \textit{L2}-norm operator.

% \textit{Assumption 2($\mu$-strongly convex):} We assume that $F (\boldsymbol{\theta})$ is strongly convex with positive parameter $\mu$, such that
% \begin{align}
% &F \left ({\boldsymbol{\theta} }\right) \geq F \left ({\boldsymbol{\theta}^{\prime} }\right) +\left ({\boldsymbol{\theta} - \boldsymbol{\theta}^{\prime}}\right)^{T} \nabla F \left ({\boldsymbol{\theta}^{\prime} }\right) \\ \notag
% &\qquad\qquad\qquad\qquad {+\,\frac {\mu } 2 \Vert \boldsymbol{\theta} - \boldsymbol{\theta}^{\prime}\Vert^2, \quad\forall\boldsymbol{\theta},\boldsymbol{\theta}^{\prime} \in \mathbb{R}^d,}
% \end{align}
% where $T$ stands for matrix transpose and $\mu$ is a positive modulus. 
\textit{Assumption 2(convex):} We assume that $F (\boldsymbol{\theta})$ is convex on $\boldsymbol{\theta}$, such that
\begin{align}
&F \left ({\boldsymbol{\theta} }\right) \geq F \left ({\boldsymbol{\theta}^{\prime} }\right) +\nabla F\left(\boldsymbol{\theta}^{\prime}\right)^\top\left(\boldsymbol{\theta}-\boldsymbol{\theta}^{\prime}\right),  \quad\forall\boldsymbol{\theta},\boldsymbol{\theta}^{\prime} \in \mathbb{R}^d,
\end{align}
where $\top$ stands for matrix transpose. 

Due to the channel errors, we assume that there is a theoretical global gradient $g^t$ obtained from the weighted average of the PS and the actual global gradient ${\widehat{g}^t}$ via the channel. With \textit{Assumption 1,2}, it is easy to get
\begin{equation}
    F\left(\boldsymbol{\theta}\right) \leq F\left(\boldsymbol{\theta}^{\prime}\right)+\nabla F\left(\boldsymbol{\theta}^{\prime}\right)^\top\left(\boldsymbol{\theta}-\boldsymbol{\theta}^{\prime}\right)+\frac{\ell}{2}\left\Vert\boldsymbol{\theta}-\boldsymbol{\theta}^{\prime} \right\Vert^2.
    \label{eq:lsmooth}
\end{equation}

Considering ${\widehat{g}^t}$ with channel errors, based on \eqref{eq:gloal update}, we can get $\boldsymbol{\theta}^{t+1} = \boldsymbol{\theta}^{t}-\eta \widehat{\boldsymbol{g}}^{t}$. Then replacing $\boldsymbol{\theta}$, $\boldsymbol{\theta}^{\prime}$ with $\boldsymbol{\theta}^{t+1}$, $\boldsymbol{\theta}^t$ and applying it into \eqref{eq:lsmooth}, we have
\begin{equation}
    F\left(\boldsymbol{\theta}^{t+1}\right) \leq F\left(\boldsymbol{\theta}^{t}\right)+ \left(\boldsymbol{g}^t\right)^\top\left(-\eta {\widehat{\boldsymbol{g}}^t} \right)+\frac{\ell \eta^2}{2}\big\Vert {\widehat{\boldsymbol{g}}^t} \big\Vert^2.
\end{equation}

Then we take expectations on both sides of the inequality, it comes that
\begin{small}
\begin{align}
    \mathbb{E}&\left\{F\left(\boldsymbol{\theta}^{t+1}\right) \right\} \notag \\ 
    &= \mathbb{E}\left\{F\left(\boldsymbol{\theta}^{t}\right)\right\} - \mathbb{E}\bigg\{\left(\eta-\ell\eta^2\right)\left(\boldsymbol{g}^t-\widehat{\boldsymbol{g}}^t \right)^{\top}\boldsymbol{g}^t \notag \\
    &\quad\quad\quad\qquad-\left(\eta-\frac{\ell \eta^2}{2}\right)\big\Vert \boldsymbol{g}^t\big\Vert^2+\frac{\ell\eta^2}{2}\big\Vert \boldsymbol{g}^t - \widehat{\boldsymbol{g}}^t\big\Vert^2  \bigg\} \notag \\
    &\overset{(a)}{=}\mathbb{E}\left\{F\left(\boldsymbol{\theta}^{t}\right)\right\} +\frac{1}{2\ell} \mathbb{E}\bigg\{\big\Vert \boldsymbol{g}^t - \widehat{\boldsymbol{g}}^t\big\Vert^2  \bigg\}-\frac{1}{2\ell} \big\Vert \boldsymbol{g}^t\big\Vert^2,
\end{align}
\end{small}where (a) is because we set the learning rate $\eta = \frac{1}{\ell}$. Then incorporating \eqref{eq:aggregation} and subtracting $\mathbb{E}\{F(\boldsymbol{\theta}^*)\}$ in both sides of the above inequality, where $\boldsymbol{\theta}^*$ is the optimal global FL model, we have the lower bound as
\begin{small}
\begin{align}
     \mathbb{E}&\left\{F\left(\boldsymbol{\theta}^{t+1}\right) - F\left(\boldsymbol{\theta}^{*}\right) \right\} \notag \\
     &\leq \mathbb{E}\left\{F\left(\boldsymbol{\theta}^{t}\right) - F\left(\boldsymbol{\theta}^{*}\right) \right\} - \frac{1}{2\ell D_{all}^2} \sum_{k=1}^{K} D_k^2\big( \big\Vert \boldsymbol{g}_k^t\big\Vert^2\notag \\
     & \qquad \qquad \qquad\qquad\qquad\qquad- \mathbb{E}\left\{\big\Vert\boldsymbol{g}_k^t-\widehat{\boldsymbol{g}}_k^t\big\Vert^2\right\} \big).
     \label{eq:loss_gap}
\end{align}
\end{small}

The above inequality indicates that the anticipated difference between the global loss function value and the optimal loss is constrained by the sum of three components. The first term on the right side of \eqref{eq:loss_gap} represents the expected difference from the prior communication round. And the second term is directly related to the squared norm of the true local gradient $\boldsymbol{g}_k^t$ with the influence of its dataset size  $D_k$. We can see that the first two terms are independent of the modulation allocating scheme. The third term is proportional to the gradient error of the aggregated global gradient, which is contingent upon the modulation order allocating design and necessitates optimization. Consequently, a reduction in the error of the aggregated global gradient will lead to a more rapid decrease in the global loss. Next, we investigate the relationship between SER and the rate of loss decay.

We first consider that only one symbol will make an error in wireless transmission and we denote ${g}_{k,z}^t$ as the theoretical local gradient of $z$-th parameter of client $k$ and $\widehat{{g}}_{k,z}^t$ as the actual global gradient via the channel. Based on the lemma in \cite{qu2024}, the expectation of $\widehat{{g}}_{k,z}^t$ is derived as
\begin{small}
\begin{align}
    &\mathbb{E} \left\{ \widehat{{g}}_{k,z}^t\right| {g}_{k,z}^t \}\notag\\
    =&  {g}_{k,z}^t+\frac{1}{\sigma}\{\underbrace{q_k(1-q_k)^{\sigma-1}\frac{range_g}{2^{\phi}-1}\sum_{i=\phi-\log_2 M}^{\phi-1}(1-2{g}_{k,z}^{t(i)})2^i}_{\textrm {bias by the first symbol error}}\notag\\
    +&\cdots+\underbrace{q_k(1-q_k)^{\sigma-1}\frac{range_g}{2^{\phi}-1}\sum_{i=0}^{\phi-(\sigma-1)\log_2 M-1}(1-2{g}_{k,z}^{t(i)})2^i}_{\textrm {bias by the last symbol error}}\} \notag\\
    =&{g}_{k,z}^t+\frac{range_g\cdot q_k(1-q_k)^{\sigma-1}}{\sigma (2^{\phi}-1)}\sum_{i=0}^{\phi-1}(1-2{g}_{k,z}^{t(i)})2^i,
    \label{eq:bias_sca}
    % \frac{1}{\sigma}\sum_{i=1}^\sigma \sum_{j=1,j\neq i}^M p_{i,j} \big\Vert s_i-\hat{s}_j\big\Vert^2 = \Delta,
\end{align}
\end{small}where $\sigma = \lceil\phi /\log_2 M_{k,n}\rceil$ represents the number of symbols contained in a parameter and each symbol contains $\log_2 M_{k,n}$ bits. For a given modulation order, $\sigma$ is fixed when characterizing the symbol-error-induced gradient distortion. $range_g \overset{\Delta}{=} \max(g)-\min(g) $ and we denote ${g}_{k,z}^{t(i)} \in \{0,1\}$ as the $i$-th digit. It is emphasized that the second component on the right of \eqref{eq:bias_sca} represents a non-zero expected bias, which arises due to the presence of random bit errors. Hence there is $\mathbb{E}\big\{(1-2{g}_{k,z}^{t(i)})^2(2^i)^2 \big\}=(2^i)^2$. Let $\epsilon$ be the bias for the $z$-th parameter of the $k$-th client, i.e., $\epsilon = {g}_{k,z}^t- \widehat{{g}}_{k,z}^t$, then we get the mean square bias as follows
\begin{align}
    \Delta \overset{\Delta}{=}\mathbb{E}\big\{\left\Vert\epsilon\right\Vert^2 \big\}= q_k(1-q_k)^{\sigma-1} \frac{range_g^2(4^\phi -1)}{3(2^\phi - 1)^2}.
\end{align}

The above formula calculates the bias in case of a single symbol error in each parameter. It is known that errors in symbols are independent of each other, with $n$ symbol errors existing, we have
\begin{small}
\begin{align}
    &\mathbb{E} \left\{\big\Vert {g}_{k,z}^t- \widehat{{g}}_{k,z}^t\big\Vert^2\right\} \notag\\
    =& \mathbb{E}\left\{\bigg\Vert\sum_{i=1}^n \epsilon_i \bigg\Vert^2\right\} \overset{(b)}{\leq} n \mathbb{E} \left\{\sum_{i=1}^n\big\Vert\epsilon_i\big\Vert^2\right\} = n^2 \Delta,
\end{align}
\end{small}where $\epsilon_i$ represents the mean square error of the $i$-th wrong symbol. The (b) is deduced from Jensen's inequality. Applying the probability of existing $n$ symbol errors in a parameter ($\sigma$ symbols), we can continue to deduce that the average error of the gradient is
\begin{small}
    \begin{align}
    &\quad\mathbb{E} \left\{\big\Vert {g}_{k,z}^t- \widehat{{g}}_{k,z}^t\big\Vert^2\right\} \notag\\
    \leq &\sum _{n=0}^{\sigma } \binom{\sigma}{n}\left ({1-q_{k}}\right)^{\sigma -n}{q_{k}}^n n^2 \Delta \notag\\
    =& \sum _{n=0}^{\sigma }\binom{\sigma}{n}\left ({1-q_{k}}\right)^{\sigma -n}{q_{k}}^n n(n-1) \Delta \notag\\
    &\qquad\qquad\qquad+\sum _{n=0}^{\sigma } \binom{\sigma}{n}\left ({1-q_{k}}\right)^{\sigma -n}{q_{k}}^n n \Delta.
\end{align}
\end{small}

We make $n(n-1)=2\binom{n}{2}$ and $n=\binom{n}{1}$. With basic computational properties of combinatorial numbers, i.e., $\binom{a}{b}\binom{b}{c}=\binom{a-c}{b-c}\binom{a}{c}$, we have
\begin{small}
    \begin{align}
    &\quad\mathbb{E} \left\{\big\Vert {g}_{k,z}^t- \widehat{{g}}_{k,z}^t\big\Vert^2\right\} \notag\\
    \leq & \sigma (\sigma -1)\Delta\sum _{n=2}^{\sigma } \binom{\sigma-2}{n-2}\left ({1-q_{k}}\right)^{\sigma -n}{q_{k}}^{n} \notag\\
    &\qquad\qquad\qquad+\sigma \Delta\sum _{n=1}^{\sigma } \binom{\sigma-1}{n-1}\left ({1-q_{k}}\right)^{\sigma -n}{q_{k}}^{n} \notag\\
    \overset{(c)}{=}&\sigma \Delta q_k \left[\left(\sigma-1\right)q_k+1 \right],
\end{align}
\end{small}where (c) is deduced from binomial theorem. Next, we conduct the calculation on the level of gradient vectors and we can get the expectation of the average gradient error for the $k$-th client as follows
\begin{small}
    \begin{align} 
    \mathbb{E}\left\{ \left\|g_k^t-\hat g_k^t\right\|^2 \right\} 
    &= \sum_z \mathbb{E}\left\{ \left\|g_{k,z}^{t}-\hat g_{k,z}^{t}\right\|^2 \right\} \nonumber\\ 
    &\le \sigma Z \frac{\mathrm{range}_g^2(4^{\phi}-1)} {3(2^{\phi}-1)^2} \left((\sigma-1)q_k^3+q_k^2\right) \nonumber\\ 
    &\le \xi \left\|g_k^t\right\|^2 \left((\sigma-1)q_k^3+q_k^2\right), 
    \label{eq:gradient_err}
\end{align}
\end{small}where $\xi= \frac{\sigma Z(4^\phi -1)}{3(2^\phi - 1)^2}$, which is determined by the quantitative bit number and  the total number of parameters. Substituting \eqref{eq:gradient_err} into \eqref{eq:loss_gap}, we have the following theorem as

\textit{Theorem 1:} Given the learning rate $\eta = \frac{1}{\ell}$, the SER $q_k$ of client $k$, the dataset size $D_{all}$ of clients participated in local training, the dataset size $D_k$ of client $k$ and the optimal global FL model $\boldsymbol{\theta}^*$,with the assumptions, the bound of  $ \mathbb{E}\left\{F\left(\boldsymbol{\theta}^{t+1}\right) - F\left(\boldsymbol{\theta}^{*}\right) \right\}$ can be given by
\begin{small}
\begin{align}
     \mathbb{E}&\left\{F\left(\boldsymbol{\theta}^{t+1}\right) - F\left(\boldsymbol{\theta}^{*}\right) \right\} \leq \mathbb{E}\left\{F\left(\boldsymbol{\theta}^{t}\right)-F\left(\boldsymbol{\theta}^{*}\right) \right\} \notag \\
     &  -\underbrace { \frac{1}{2\ell D_{all}^2} \sum_{k=1}^{K} \left(D_k \big\Vert \boldsymbol{g}_{k}^t\big\Vert\right)^2 \left\{ 1 - \xi\left[\left(\sigma-1\right)q_k^3+q_k^2 \right] \right\}}_{\textrm {Impact of wireless factors on FL convergence, defined as $\delta$ in III.B}}.
     \label{eq:theorem}
\end{align}
\end{small}

Compared to \eqref{eq:loss_gap}, we can see that the annotated term, the gradient error of the aggregated global gradient, is denoted by $q_k$. Specifically, an increase in $q_k$ leads to a greater discrepancy between the theoretical global gradient $\boldsymbol{g}^t$ and the actual global gradient $\widehat{\boldsymbol{g}}^t$, leading to a reduction in the efficacy of loss decay. Therefore, a higher modulation order may reduce uplink latency but can also increase gradient distortion and slow down FL convergence. 

This result provides the analytical basis for the convergence-latency aware adaptive modulation and resource allocation design in the following subsection. It should be noted that the bound is not claimed to prove convergence for non-convex neural networks; rather, it motivates a surrogate communication metric that penalises SER-induced gradient distortion. For non-convex neural networks, the bound should be interpreted as indicating the relative impact of modulation-induced gradient distortion rather than as a strict convergence guarantee.

\subsection{Problem Formulation}
From \eqref{eq:theorem}, we have seen that the efficacy of loss decay is mainly related to client gradient modulus, client data volume and SER. Therefore, we define the loss decay as $\delta$. The larger $\delta$ is, the faster the loss function decreases, which is capable of assessing the rate of convergence while ensuring the minimal decay of global loss. Moreover, there is a positive correlation among $D_k \Vert \boldsymbol{g}_k^t \Vert^2$ and $\delta$, which is in line with common sense, i.e., a larger data volume and a greater gradient value contribute more significantly to the model update. In our \textit{Section II.C}, we have presented the theory that the modulation order has an impact on both the transmission rate and SER. A higher modulation order can achieve a higher data transmission rate, but it also leads to greater signal distortion and a higher SER. Allocating different sub-channels to clients can result in varying SER and transmission delay. Therefore, we construct an objective function with the aim of achieving a faster model convergence rate with the smallest possible delay, thereby ensuring the effectiveness of the model. In each communication round, we denote
\begin{align}
    % P = \frac{\frac{1}{2\ell D_{all}^2} \sum_{k=1}^{K} \left(D_k \big\Vert \boldsymbol{g}_{k}^t\big\Vert\right)^2 \left\{ 1 - \sigma Z \Delta\left[\left(\sigma-1\right)q_k^2+q_k \right] \right\}}{\max_k\{T_k^C + T_k^U\}+T^D}
    obj =& \delta -\lambda T \notag\\ 
    =& \frac{1}{2\ell D_{all}^2} \sum_{k=1}^{K} \left(D_k \big\Vert \boldsymbol{g}_{k}^t\big\Vert\right)^2 \left\{ 1 - \xi\left[\left(\sigma-1\right)q_k^3+q_k^2 \right] \right\} \notag \\ 
    & - \lambda\left\{\max_k\{T_k^C + T_k^U\}+T^D \right\},
    \label{eq:obj}
\end{align}
where $\lambda \in [0,\lambda_{max})$ is a minimal weight coefficient,  which serves to slightly punish the delay, ensuring that the solution with a smaller delay is preferred among solutions with similar convergence speeds. Consequently, the joint optimization of modulation selection and resource allocation is imperative, with the enhancement of \eqref{eq:obj} serving as the key design objective. 

According to what we described in \textit{Section II}, the interplay between modulation selection and spectrum allocation has been identified as a critical factor influencing the key performance metrics of our FL system. To address these technical challenges, our optimization strategy focuses on identifying optimised combinations of modulation schemes and spectrum resource allocation patterns that improves the convergence-latency objective, particularly under the constraints of unreliable wireless communication channels. The fundamental optimization challenge lies in achieving an optimal balance between system convergence rate and learning latency, which is formulated as
\begin{small}
\begin{align}
    \mathcal{P}: \max_{\boldsymbol{a}_k, \boldsymbol{M}_k} &obj=\delta -\lambda T ,\label{opt}\\
    \mathrm{s.t.}\quad
    &a_{k,n}\in \{0,1\}, \forall k \in \mathcal{K}, \forall n ,\notag\tag{36a}\\
    &\sum _{k=1}^{K}a_{k,n}\leq 1, \forall n ,\notag\tag{36b}\\
    &\sum_{n=1}^{N}a_{k,n}\ge 1,\quad \forall k\in\mathcal{K},\notag\tag{36c}\\
    &q_{k}\leq q_{\max },\forall k \in \mathcal {K},\notag\tag{36d}\\
    &M_{k,n}=2^{j}, j \in \mathbb{Z}_{> 0}  |(MPSK),\notag\tag{36e}\label{35d}\\
    &M_{k,n}=4^{j}, j \in \mathbb{Z}_{> 0}  |(MQAM),\notag\tag{36f}\label{35e}
\end{align}
\end{small}

\textit{Remark 1:}
The SER threshold $q_{\max}$ in (36d) is selected to ensure a positive
convergence-related loss decay. According to Theorem 1, it is sufficient
to choose $q_{\max}$ such that $1-\xi\left((\sigma_{\max}-1)q_{\max}^3+q_{\max}^2\right)>0$, where $\sigma_{\max}$ denotes the largest possible value of
$\sigma_{k,n}$ over the considered modulation set. Under this
condition, the convergence-related term in Theorem 1 remains
positive for all feasible clients.

% This condition guarantees that the wireless-error-related factor in
% Theorem 1 remains positive for all feasible modulation orders.
The problem can be characterized as a discrete optimization problem, which involves the joint optimization of wireless resource allocation ($a_{k,n}$) and modulation order ($M_{k,n}$). In the following section, we will specifically elaborate on how we will solve the optimization problem \eqref{opt}.

\section{Problem Optimization}

The optimization problem is a mixed-integer nonlinear programming (MINLP) problem due to the binary sub-channel allocation variables, discrete modulation orders, nonlinear SER functions, and the max-latency term. To obtain a tractable solution, we adopt a hybrid alternating optimization framework, where different optimization techniques are employed for different subproblems. Specifically, given the sub-channel allocation, the modulation-order subproblem is solved via continuous relaxation and Newton iterations with discrete projection. Given the modulation order, the sub-channel allocation is optimized through binary relaxation followed by KKT-based optimization. The two subproblems are solved alternately until convergence.

\subsection{Modulation Order Optimization}
% The modulation order plays a crucial role in determining the trade-off between communication efficiency and system performance. The goal of modulation order optimization is to select the modulation scheme that maximizes the convergence speed of the model while minimizing the transmission latency, all under the constraints imposed by the communication environment.

The objective function for modulation order optimization is given by
\begin{align}
    obj\left(M_{k,n}\right) = \delta\left(M_{k,n}\right)-\lambda T\left(M_{k,n}\right),
\end{align}
where $\delta\left(M_{k,n}\right)$ is the loss decay function, reflecting the impact of gradient errors on model convergence, and $T\left(M_{k,n}\right)$ is the communication latency due to the modulation scheme. To simplify the analysis of communication error impact, we consider the uplink SER expression, which is explained in Appendix A.

Since the modulation order is selected from a finite discrete set, the modulation order optimization subproblem is combinatorial in nature. To obtain a low-complexity solution, we first solve a continuous relaxation, where $M_{k,n}$ is allowed to vary within $[M_{\min},M_{\max}]$. For the relaxed problem, the first-order stationary condition is given by
\begin{equation}
    f(M_{k,n}) \overset{\Delta}{=} \frac{\partial obj(M_{k,n})}{\partial M_{k,n}}=0.
\end{equation}

The continuous solution is obtained by the Newton-Raphson method as
\begin{equation}
M_{k,n}^{(i+1)}
=
\Pi_{[M_{\min},M_{\max}]}
\left(
M_{k,n}^{(i)}
-
\frac{f(M_{k,n}^{(i)})}{f'(M_{k,n}^{(i)})}
\right),
\end{equation}
where $f'(M_{k,n})$ denotes the derivative of $f(M_{k,n})$, and $\Pi_{[M_{\min},M_{\max}]}(\cdot)$ is the projection operator that guarantees the feasibility of the relaxed solution. The obtained solution is regarded as a stationary point of the relaxed modulation subproblem. Since $\sigma_{k,n}=\lceil \phi/\log_2(M_{k,n})\rceil$ is a piecewise constant function, it is treated as locally fixed during each Newton update and recalculated after each iteration.

After convergence, the obtained stationary solution $\widetilde{M}_{k,n}$ is projected onto the discrete modulation set. To avoid infeasible high-order modulation, we do not simply select the nearest discrete value. Instead, the neighboring discrete modulation orders around $\widetilde{M}_{k,n}$ are evaluated, and the final modulation order is selected as
\begin{equation}
M_{k,n}^{\star}
=
\arg\max_{M\in \mathcal{M}_{k,n}^{\rm feas}\cap \mathcal{N}(\widetilde{M}_{k,n})}
obj(M),
\end{equation}
where $\mathcal{N}(\widetilde{M}_{k,n})$ denotes the neighboring discrete modulation orders of $\widetilde{M}_{k,n}$, and
\begin{equation}
\mathcal{M}_{k,n}^{\rm feas}
=
\{M\in\mathcal{M}: q_{k,n}(M)\le q_{\max}\}.
\end{equation}

If no feasible modulation order exists in $\mathcal{N}(\widetilde{M}_{k,n})$, the modulation order is gradually reduced until the SER constraint is satisfied.

\subsection{Sub-Channel Allocation Optimization}

Given the optimized modulation orders, we then optimize the sub-channel allocation policy. Since the original sub-channel allocation variable $a_{k,n}$ is binary, the resulting problem is a combinatorial optimization problem. To obtain a tractable solution, we first relax the binary constraint as $0\le a_{k,n}\le 1,\quad \forall k\in\mathcal{K},\ \forall n\in\mathcal{N}$.

After solving the relaxed problem, a binary recovery operation is performed to obtain the final sub-channel allocation.

For notational simplicity, we define the convergence-related utility of allocating sub-channel $n$ to client $k$ as
\begin{align}
\alpha_{k,n}
=
\frac{D_k^2\|g_k^t\|^2}{2\ell D_{\rm all}^2}
\left[
1-\xi\left((\sigma_{k,n}-1)q_{k,n}^3+q_{k,n}^2\right)
\right],
\end{align}
where $q_{k,n}$ denotes the SER of client $k$ on sub-channel $n$ under the selected modulation order, and $\sigma_{k,n}=\lceil \phi/\log_2(M_{k,n})\rceil$. Moreover, the effective uplink transmission rate of client $k$ is given by
\begin{align}
R_k(\mathbf{a}_k)
=
\sum_{n=1}^{N}a_{k,n}B_n\log_2(M_{k,n}).
\end{align}

Accordingly, the uplink latency of client $k$ can be rewritten as
\begin{align}
T_k^U(\mathbf{a}_k)
=
\frac{\phi Z}{R_k(\mathbf{a}_k)}.
\end{align}

To handle the non-smooth maximum latency term, we introduce an auxiliary variable $\tau$ satisfying
\begin{align}
T_k^C+\frac{\phi Z}{R_k(\mathbf{a}_k)} \le \tau,\quad \forall k\in\mathcal{K}.
\end{align}

Then, the relaxed sub-channel allocation problem can be formulated as
\begin{small}
\begin{align}
\textbf{P}_{\rm RA}: \quad
\max_{\mathbf{a},\tau}\quad
& \sum_{k=1}^{K}\sum_{n=1}^{N}a_{k,n}\alpha_{k,n}
-\lambda \tau \\
\mathrm{s.t.}\quad
& T_k^C+\frac{\phi Z}
{\sum_{n=1}^{N}a_{k,n}B_n\log_2(M_{k,n})}
\le \tau,\quad \forall k,\notag \tag{46a}\\
& \sum_{k=1}^{K}a_{k,n}\le 1,\quad \forall n,\notag \tag{46b}\\
& \sum_{n=1}^{N}a_{k,n}\ge 1,\quad \forall k,\notag \tag{46c}\\
& 0\le a_{k,n}\le 1,\quad \forall k,n.\notag \tag{46d}
\end{align}
\end{small}

Problem $\textbf{P}_{\rm RA}$ is a continuous relaxation of the original sub-channel allocation problem. Compared with the original formulation, the auxiliary variable $\tau$ transforms the maximum latency term into a set of smooth inequality constraints, which enables the application of KKT-based analysis.

The optimality conditions of $\textbf{P}_{\rm RA}$ are derived in Appendix B. Based on the obtained relaxed solution $\tilde a_{k,n}$, we recover the binary allocation by assigning each sub-channel to the client with the largest relaxed allocation value, i.e.,
\begin{align}
a_{k,n}^{\star}
=
\begin{cases}
1, & k=\arg\max\limits_{j\in\mathcal{K}}\tilde a_{j,n},\\
0, & \mathrm{otherwise}.
\end{cases}
\end{align}
% After the binary recovery, the SER constraint is checked for each client. If the recovered allocation violates the SER constraint, the corresponding sub-channel is reassigned to the next feasible client or the modulation order is reduced according to the modulation order optimization procedure. It is worth noting that the KKT-based solution characterizes the optimality of the relaxed sub-channel allocation problem. Due to the binary recovery step, the obtained solution is a low-complexity feasible solution to the original mixed-integer problem rather than a globally optimal solution. 
After binary recovery, the SER constraint is checked for each client, and infeasible assignments are corrected by sub-channel reassignment or modulation-order reduction. Since the KKT analysis is performed for the relaxed problem, the recovered binary solution is a low-complexity feasible solution rather than a globally optimal solution to the original mixed-integer problem.

Algorithm 1 summarizes the overall procedure, whose complexity is $O(I(KNR_N+C_{\rm RA}))$, where $I$, $R_N$, and $C_{\rm RA}$ denote the numbers of alternating iterations, Newton iterations, and the complexity of the relaxed allocation subproblem, respectively. 
% This complexity is substantially lower than that of exhaustive search over all modulation and allocation combinations.

\begin{algorithm}[t]
\caption{Joint Adaptive Modulation and Sub-Channel Allocation}
\label{alg:joint}
\begin{algorithmic}[1]
\REQUIRE Client set $\mathcal{K}$, sub-channel set $\mathcal{N}$, channel information, dataset sizes $\{D_k\}$, local gradient norms $\{\|g_k^t\|\}$, modulation set $\mathcal{M}$, SER threshold $q_{\max}$, trade-off parameter $\lambda$, maximum iteration number $I_{\max}$, and convergence tolerance $\epsilon$.
\ENSURE Sub-channel allocation $\mathbf{a}^{\star}$ and modulation order $\mathbf{M}^{\star}$.

\STATE Initialize a feasible sub-channel allocation $\mathbf{a}^{(0)}$ and set $i=0$.
\STATE Compute the initial objective value $obj^{(0)}$.

\REPEAT
    \STATE Given $\mathbf{a}^{(i)}$, obtain the modulation order $\mathbf{M}^{(i+1)}$ according to the modulation order optimization method in Section~IV-A.

    \STATE Given $M^{(i+1)}$, solve the relaxed sub-channel allocation problem in Section IV-B.

    \STATE Recover $a^{(i+1)}$ and correct infeasible assignments to satisfy the allocation and SER constraints.

    \STATE Compute the objective value $obj^{(i+1)}$ according to \eqref{eq:obj}.
    % \[
    % Obj^{(i+1)}
    % =
    % \Psi(\mathbf{a}^{(i+1)},\mathbf{M}^{(i+1)})
    % -
    % \lambda T(\mathbf{a}^{(i+1)},\mathbf{M}^{(i+1)}).
    % \]

    \STATE Set $i=i+1$.

\UNTIL{$|obj^{(i)}-obj^{(i-1)}|\le \epsilon$ or $i\ge I_{\max}$}

\STATE Set $\mathbf{a}^{\star}=\mathbf{a}^{(i)}$ and $\mathbf{M}^{\star}=\mathbf{M}^{(i)}$.
\RETURN $\mathbf{a}^{\star}$ and $\mathbf{M}^{\star}$.
\end{algorithmic}
\end{algorithm}

\section{Experimental Results}
In this section, we evaluate the performance of the proposed adaptive modulation and resource allocation scheme through extensive simulations. The proposed method is compared with several benchmark schemes to demonstrate its effectiveness.
\subsection{Experiment Setup}
We consider a RIS-assisted wireless federated learning system consisting of one parameter server and \(K\) edge clients. The system bandwidth is divided into \(N\) orthogonal sub-channels using OFDMA. We set the channel path loss exponent model to PL, where $PL[dB]=128.1+37.6\log(d)$ with $d$ representing the distance between clients and the PS in kilometer. The LoS condition is modeled as a distance-dependent random process following the 3GPP TR 36.828 specification\cite{3gpp_tr36_828_short}. For the link with distance $d$, the LoS probability is given by $P_{\mathrm{LoS}}(d) = \min\!\left(\frac{a}{d},\,1\right)\left(1 - e^{-d/b}\right) + e^{-d/b}$, where $a$ and $b$ are scenario-dependent constants. Unless otherwise specified, the remaining simulation parameters are summarized in Table~\ref{tab:sim_params}.

\begin{table}[t]
\caption{Main Simulation Parameters}
\label{tab:sim_params}
\centering
\begin{tabular}{lc}
\hline
\textbf{Parameter} & \textbf{Value} \\
\hline
Total bandwidth \(B\) & \(10\) MHz \\
Number of RIS reflecting elements \(R\) & \(16\) \\
Noise power spectral density \(N_0\) & \(-174\) dBm/Hz \\
Path-loss model \(PL[dB]\) & \(128.1 + 37.6\log(d)\) \\
Trade-off parameter \(\lambda\) & \(10^{-1}\) \\
Number of clients \(K\) & \(\{10,20,30,40,50\}\) \\
Average SNR (dB) & \(\{10,15,20,25\}\) \\
Dirichlet parameter \(\rho\) & \(0.6\)\\
Average number of baseline experiments & \(5\)\\
\hline
\end{tabular}
\end{table}

The local datasets are uniformly and randomly partitioned among participating clients, resulting in IID and Non-IID data distribution. We examine the result of our proposed algorithm, using the following neural networks and datasets.
\begin{itemize}
\item[$\bullet$] \textbf{MLP with MNIST.} The MNIST consists of 70,000 grayscale handwritten digit images of size $28\times28$, including 60,000 training samples and 10,000 test samples across 10 classes (digits 0–9). 
% Due to its simplicity and well-balanced data distribution, MNIST is widely used as a benchmark dataset for evaluating classification performance and learning efficiency.
\item[$\bullet$] \textbf{CNN with CIFAR-10.} The CIFAR-10 contains 60,000 color images of size $32\times32$ from 10 object categories, with 50,000 training samples and 10,000 test samples. Each image consists of three RGB channels. 
% CIFAR-10 is commonly adopted to evaluate the performance of convolutional neural networks on small-scale natural image classification tasks with higher complexity than MNIST.
\item[$\bullet$] \textbf{CNN with Speech Commands.} The Google Speech Commands Dataset consists of one-second audio recordings of spoken keywords sampled at 16 kHz, collected from a large number of speakers. We adopt a standard 12-class classification setting, including 10 target commands along with additional “unknown” and “silence” classes. 
% This dataset is widely used for keyword spotting tasks and provides a more realistic evaluation scenario due to its inherent variability in speaker characteristics and background noise.
\end{itemize}

We additionally consider a Dirichlet non-IID partition on CIFAR-10. Specifically, the class proportions of each client are sampled from a Dirichlet distribution with concentration parameter $\rho=0.6$. In the following experiment, the comparison between Proposed (adaptive modulation-only scheme) and Proposed+resource allocation (RA) serves as an ablation study to evaluate the additional gain brought by resource allocation. Also, to evaluate the performance, we set some baselines to compare with the proposed scheme:
\begin{itemize}
\item[$\bullet$] \textbf{BPSK, QPSK, 16QAM.} Fixed-order modulation method.
\item[$\bullet$] \textbf{SDPR\cite{sun2025}.} A SINR-based selection mechanism, SDPR, which selects clients based on their SINRs
to allow more clients with minor errors to participate in FL.
\item[$\bullet$] \textbf{No-RIS.} The RIS is removed from the considered system, and blocked or weak NLoS links suffer from more severe attenuation and higher transmission errors, leading to degraded FL performance.
\end{itemize}
\subsection{Cross-Dataset Performance Analysis}
To comprehensively evaluate the effectiveness of the proposed scheme, we further compare its performance across multiple learning tasks with different levels of complexity, including MNIST, CIFAR-10, and Speech Commands. 
% These datasets span from simple image classification to more complex visual recognition and temporal speech understanding, thereby providing a holistic assessment of robustness under heterogeneous learning scenarios.

\begin{figure}[!t]
\centering
\includegraphics[width=2.5in]{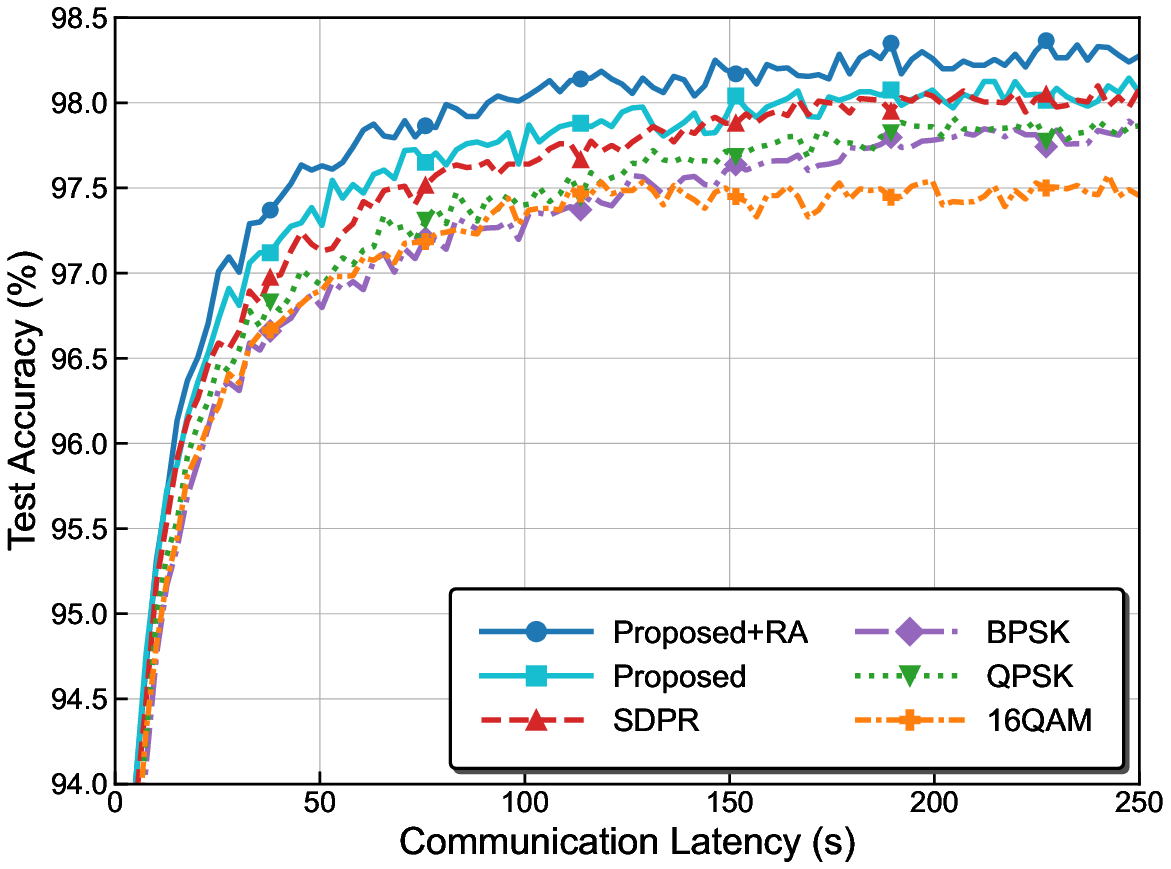}%
\caption{Performance on MNIST.}
\label{mnist}
\end{figure}

\begin{figure}[!t]
    \centering
    \subfloat[IID setting]{\includegraphics[width=2.5in]{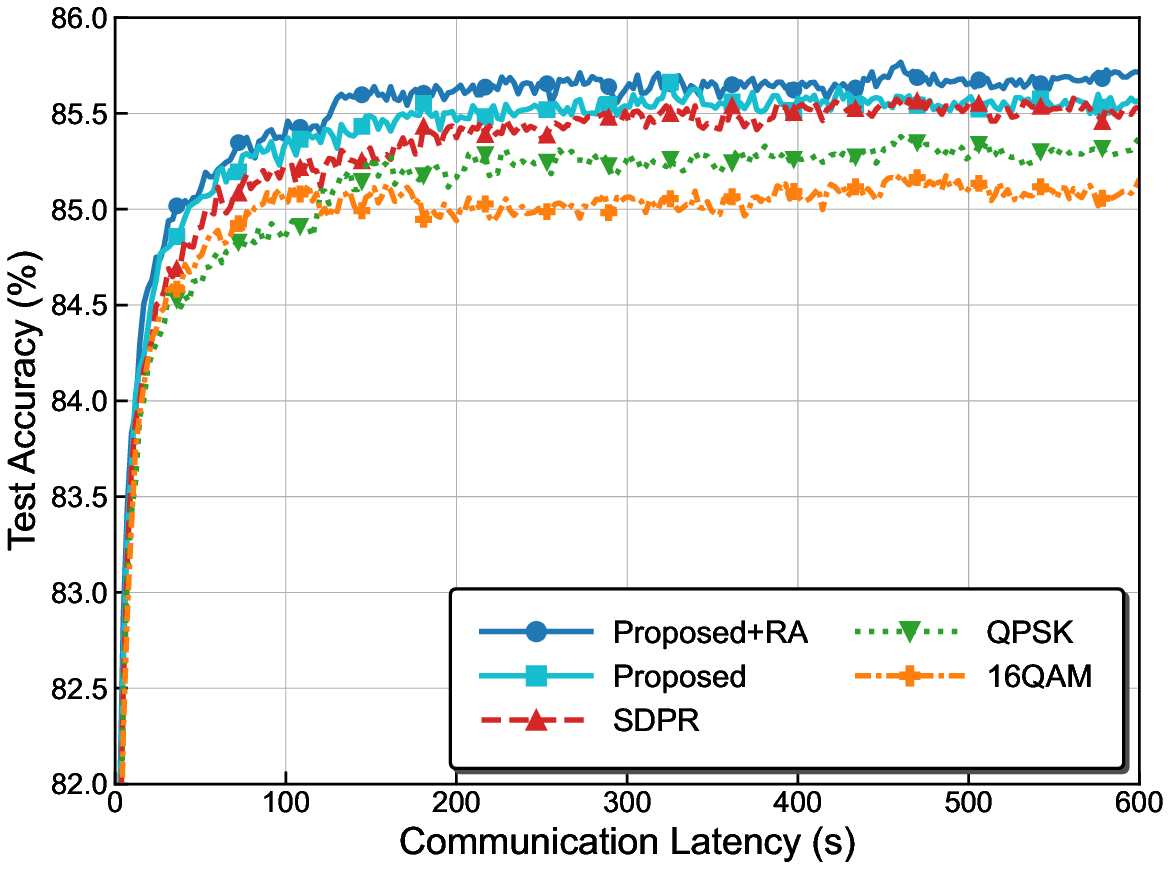}}%
    \label{cifar_iid}
    \hfil
    \subfloat[Dirichlet Non-IID setting]{\includegraphics[width=2.5in]{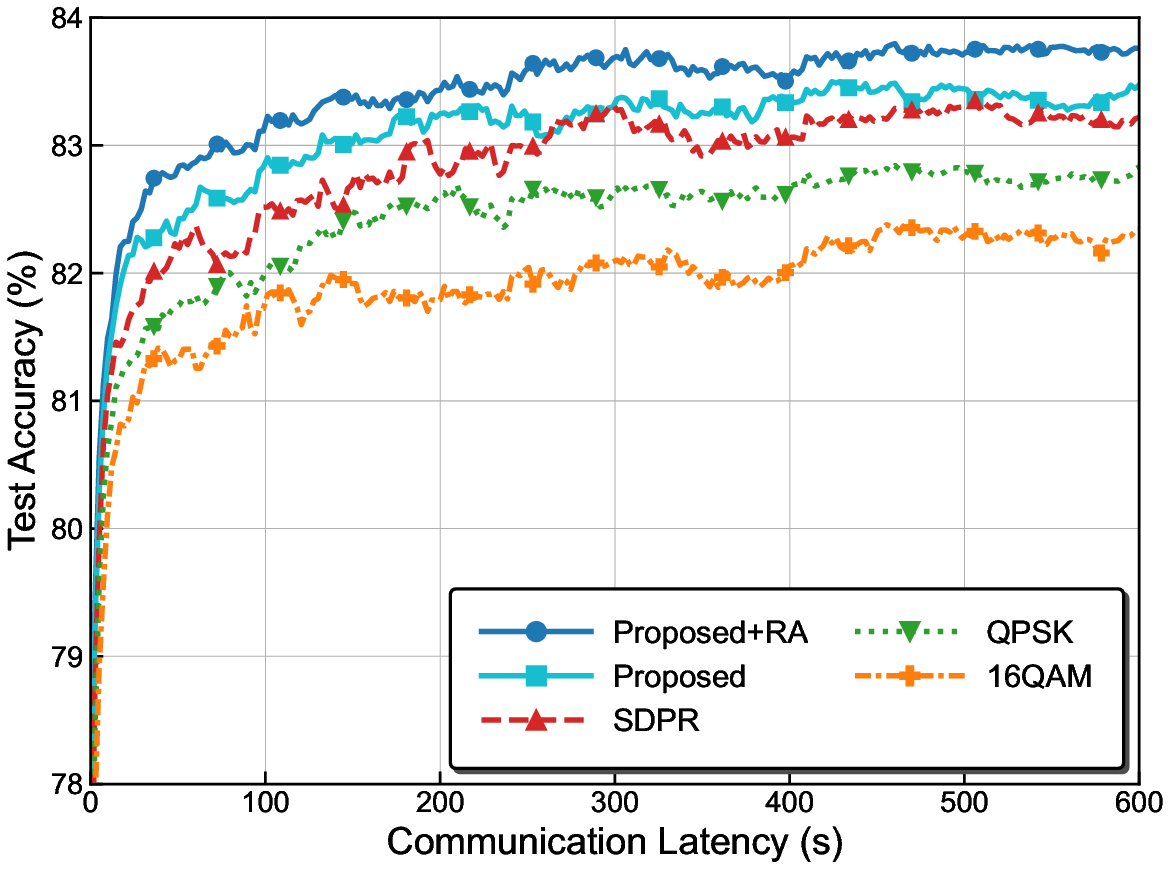}}%
    \label{cifar_noniid}
    \hfil
    \caption{Performance on CIFAR-10 under different data distributions.}
    \label{best}
\end{figure}

\begin{figure}[!t]
\centering
\includegraphics[width=2.5in]{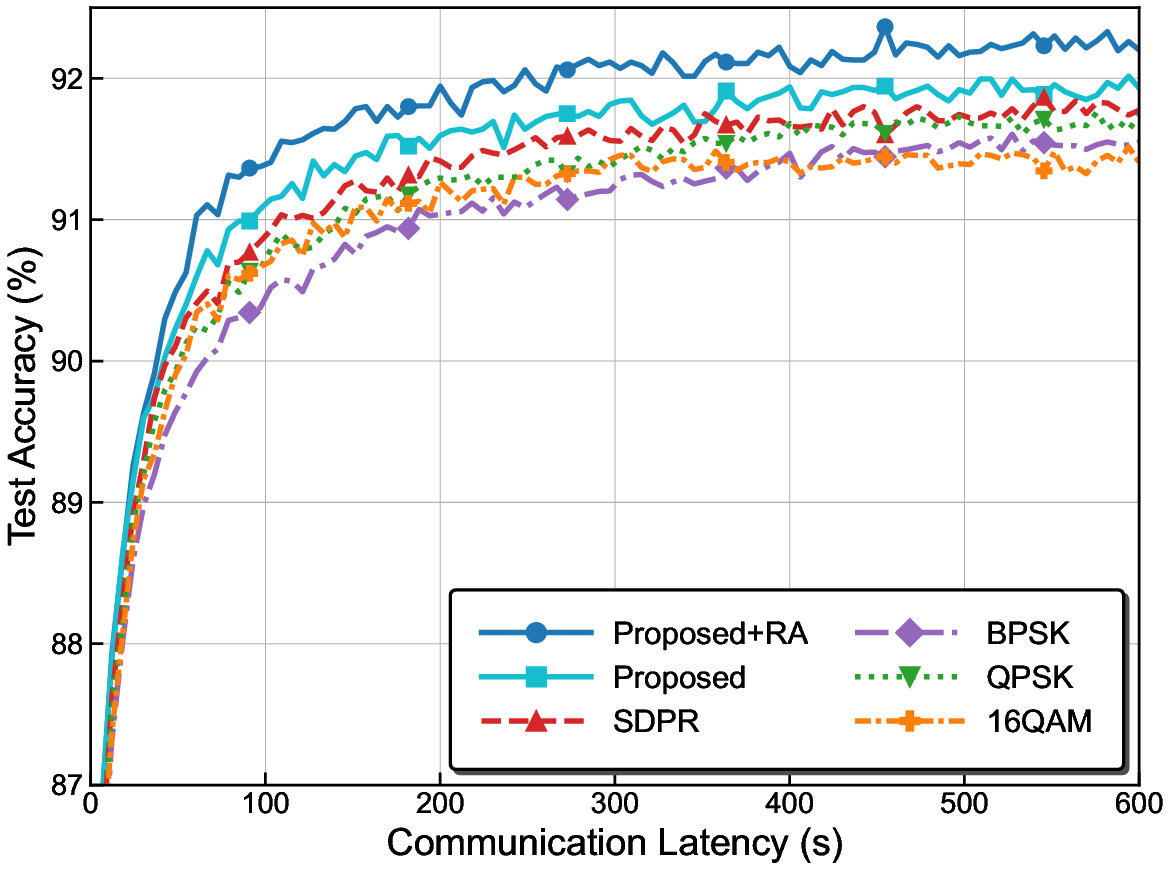}%
\caption{Performance on Speech Commands.}
\label{speech}
\end{figure}

From the results in Fig.~\ref{mnist}, Fig.~\ref{best}, Fig.~\ref{speech}, several important observations can be drawn. First, all schemes exhibit a similar convergence pattern across different datasets, where the test accuracy increases rapidly at the early stage and gradually saturates as the communication latency accumulates. For example, on the MNIST dataset, all schemes reach above 95\% accuracy within approximately 20 s, while on CIFAR-10 and Speech Commands, a greater latency is required to achieve stable convergence due to increased task complexity.

Second, as the task becomes more complex, the performance gap among different schemes becomes more evident. Taking the IID data as an example, on MNIST, the final accuracy gap between the Proposed+RA scheme and the best baseline is relatively small (around 0.3\%). In contrast, on CIFAR-10, the Proposed+RA scheme achieves approximately 85.8\%, outperforming the SDPR\cite{sun2025} method (around 85.5\%) and fixed-modulation schemes (around 85.1\%) by a noticeable margin. This gap further enlarges in the Speech Commands task, where Proposed+RA reaches about 92.5\%, compared with 92\% for the Proposed scheme, 91.5\% for SDPR\cite{sun2025}, 91.4\% for QPSK and 91\% for 16QAM.

Third, fixed-modulation schemes exhibit distinct limitations under complex tasks. Although QPSK and 16QAM achieve competitive performance on MNIST, their degradation becomes more obvious on CIFAR-10 and Speech Commands. In particular, 16QAM consistently yields the lowest accuracy due to its higher symbol error rate, while QPSK provides better reliability but lower spectral efficiency. For instance, on the Speech Commands, the accuracy gap between QPSK and 16QAM is about 0.4\%, reflecting the trade-off between reliability and transmission rate. Also, in Fig.~\ref{best}, compared with the IID setting, all schemes exhibit slower convergence and lower test accuracy under the Non-IID setting. This is because heterogeneous local data distributions increase the inconsistency among local updates, which makes global aggregation more sensitive to both gradient bias and transmission errors. As a result, the performance gap among different communication strategies becomes more pronounced.

In contrast, the proposed adaptive modulation scheme consistently outperforms all fixed-modulation baselines across all datasets. The advantage becomes increasingly significant as task complexity grows. For example, the performance gain of the Proposed scheme over fixed-modulation baselines is marginal on MNIST but increases to approximately 0.5\% on CIFAR-10 and further to around 1\% on Speech Commands. This trend indicates that adaptive modulation is particularly effective in scenarios where gradient distortion has a stronger impact on model convergence. Moreover, Proposed+RA further improves over the adaptive modulation-only scheme by about 0.3\% on CIFAR-10 and 0.5\% on Speech Commands, indicating that resource allocation becomes more beneficial when task complexity and communication contention increase. These results are consistent with Theorem 1, since reducing SER-induced gradient distortion improves the convergence-related loss decay factor. 

% Furthermore, the proposed scheme provides additional performance improvements over the adaptive modulation-only scheme. While the improvement is relatively small on MNIST, it becomes more pronounced on CIFAR-10 (about 0.2\%–0.3\%) and Speech Commands (about 0.3\%–0.5\%). This demonstrates that resource allocation plays a more critical role in complex tasks, where communication contention and latency become dominant factors affecting learning performance.

% Finally, all experimental results are well aligned with the theoretical analysis presented in Section III. As indicated by Theorem 1, the convergence behavior is closely related to the symbol error rate through the loss decay factor $\delta$. By effectively reducing gradient distortion and improving transmission efficiency, the proposed scheme achieves faster convergence and higher final accuracy across diverse learning tasks, thereby validating its robustness and scalability in practical wireless federated learning environments.

% \begin{figure}[!t]
% \centering
% \includegraphics[width=0.45\textwidth]{simulation/FL_Accuracy_vs_Latency_cifar.eps}%
% \caption{Performance under CIFAR-10 on IID dataset}
% \label{cifar}
% \end{figure}

\begin{figure}[!t]
\centering
\includegraphics[width=2.5in]{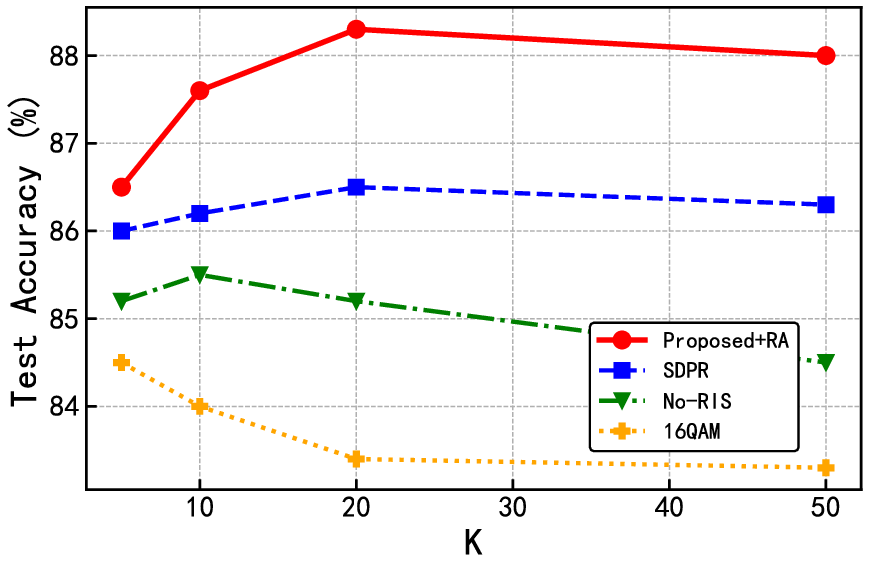}%
\caption{Performance versus the number of clients.}
\label{client}
\end{figure}

\begin{figure}[!t]
\centering
\includegraphics[width=2.5in]{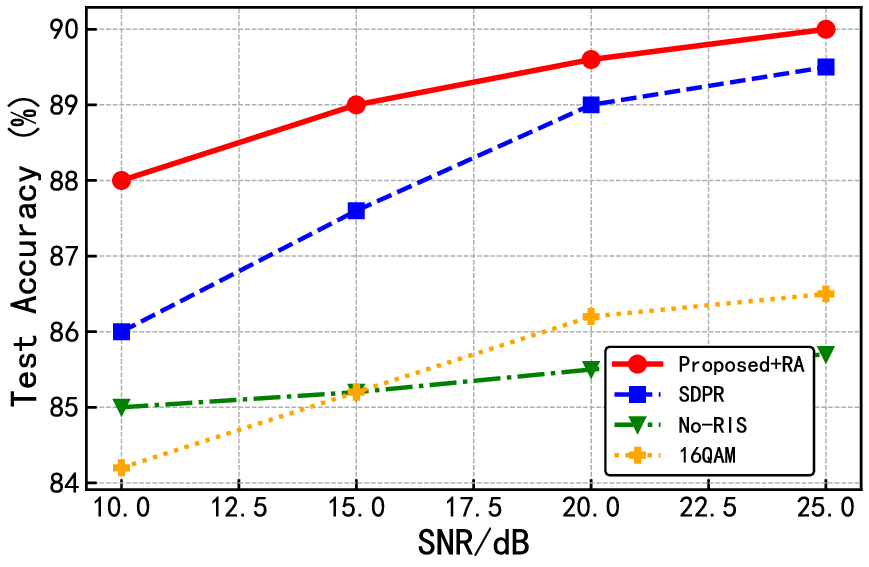}%
\caption{Performance versus the average SNR.}
\label{snr}
\end{figure}
\subsection{System-Level Performance Analysis}
To further evaluate the performance of our proposed scheme under dynamic wireless environments, we investigate its behavior with respect to the number of participating clients and the channel quality (SNR), as shown in Fig.~\ref{client} and Fig.~\ref{snr}.

Fig.~\ref{client} illustrates the impact of the number of participating clients on learning performance. Increasing the number of clients has two opposite effects: it provides more training data diversity and improves aggregation quality, but also introduces heavier communication load and more severe resource contention. The Proposed+RA scheme consistently achieves the best performance, with the accuracy increasing from about 86.5\% at \(K=10\) to 88.2\% at \(K=20\), and remains around 88\% at \(K=50\), while SDPR\cite{sun2025} stays around 86.3\%, the No-RIS baseline decreases from 85\% to 84.5\%, and fixed 16QAM drops from 84.3\% to 83.3\%. This demonstrates the superior scalability and error tolerance of our proposed scheme. In addition, the No-RIS baseline degrades more obviously as the client number increases, since more blocked or weak links lead to more severe error accumulation.

Fig.~\ref{snr} shows the learning performance under different average SNR conditions. As the SNR increases, all schemes improve because better channel quality reduces symbol errors in uplink model transmission. However, the Proposed+RA scheme remains the best-performing method over the entire SNR range. At \(10\)~dB, it achieves an accuracy of about 88\%, compared with 86\% for SDPR\cite{sun2025}, 85\% for No-RIS, and 84\% for fixed 16QAM. At \(25\)~dB, its accuracy further improves to nearly 90\%, while SDPR\cite{sun2025} reaches about 89.5\%, and the No-RIS and 16QAM baselines remain lower at around 85.7\% and 86.5\%, respectively. It is also observed that SDPR\cite{sun2025} gradually approaches the Proposed+RA as the SNR increases. This is because SDPR\cite{sun2025} relies on SNR-based client selection and RIS-assisted communication enhancement. Under better channel conditions, more clients satisfy the requirement and participate in aggregation with fewer transmission errors. Nevertheless, our proposed method still maintains a consistent advantage by jointly optimizing adaptive modulation and resource allocation, which enables a better reliability-latency trade-off and more efficient use of wireless resources.

\section{Conclusion}

This paper studied convergence-latency aware adaptive modulation and sub-channel allocation for wireless FL under direct LoS and RIS-assisted blocked NLoS transmission scenarios. By modeling local computation and communication latency, we established a convergence-related bound that explicitly captures the effect of modulation-dependent symbol errors on FL loss decay. Based on this result, we formulated a joint convergence-latency optimization problem and solved it using a low-complexity hybrid alternating optimization framework.

Simulation results on MNIST, CIFAR-10, and Speech Commands showed that our proposed scheme consistently outperforms benchmark methods in both convergence speed and final accuracy. The performance gain is particularly significant in complex tasks, dense-client settings, and low-SNR scenarios, where transmission errors and latency become the main bottlenecks. These results confirm the effectiveness of jointly optimizing communication reliability and resource efficiency for wireless FL.

\appendices
\section*{Appendix A: Detailed Derivation of Modulation Order Optimization}

\subsection*{A.1 Problem Formulation}
We begin with the optimization problem of modulation order \( M_{k,n} \) defined as
\begin{equation}
\max_{M_{k,n}} \delta(M_{k,n}) - \lambda T(M_{k,n}).
\end{equation}

The objective function can be explicitly written as
\begin{small}
\begin{align}
obj =& \frac{1}{2\ell D_{all}^2} \sum_{k=1}^{K} D_k^2 \|g_k^t\|^2 
\left[1 - \xi \left((\sigma - 1) q_{k,n}^3 + q_{k,n}^2\right)\right] 
\notag\\
&- \lambda \left(\frac{\phi Z}{B_n \log_2(M_{k,n})}\right).
\end{align}
\end{small}

\subsection*{A.2 SER Approximation and \( G(M_{k,n}) \) Definition}

The SER for \textit{M}-ary modulation is often expressed in terms of the Q-function, which involves an integral form that complicates gradient-based optimization. To circumvent this issue, we approximate the Q-function using the following exponential expression\cite{zhu2023}
\begin{align}
    Q(x) \approx \frac{1}{12}exp\left(-\frac{x^2}{2}\right)+\frac{1}{4} exp\left(-\frac{2 x^2}{3}\right),
\end{align}
which provides a tight and differentiable approximation suitable for optimization purposes.
Based on this approximation, taking the LoS scene as an example, the SER can be rewritten as a function of $M_{k,n}$
\begin{equation}
q_{k,n}(M_{k,n}) \approx \frac{1}{6}e^{-G(M_{k,n})} + \frac{1}{2}e^{-\frac{4}{3}G(M_{k,n})},
\end{equation}

where the function \( G(M_{k,n}) \) is defined as
\begin{equation}
G(M_{k,n}) = \frac{h_{k,n}^2 E_{k,n} \log_2(M_{k,n}) \sin^2\left(\frac{\pi}{M_{k,n}}\right)}{N_0}.
\end{equation}

The following derivation uses the LoS-MPSK case as an example. For other channel conditions and modulation formats, the same procedure applies by replacing $q_{k,n}(M_{k,n})$ with the corresponding SER expression in Section II-C.
\subsection*{A.3 Objective Function Differentiation}
We begin by calculating the derivative of the objective function with respect to \( M_{k,n} \)
\begin{small}
\begin{align}
\frac{\partial \text{obj}}{\partial M_{k,n}} 
=& -\frac{D_k^2 \|g_k^t\|^2 \xi}{2\ell D_{all}^2} \left[3(\sigma - 1)q_{k,n}^2 + 2q_{k,n}\right]\frac{\partial q_{k,n}}{\partial M_{k,n}} \notag \\
&- \lambda \frac{\partial T}{\partial M_{k,n}} = 0.
\end{align}
\end{small}

\subsection*{A.4 Derivation of \( \frac{\partial q_{k,n}}{\partial M_{k,n}} \) and \( \frac{\partial T}{\partial M_{k,n}} \)}
Substituting the approximation
\begin{small}
\begin{equation}
\frac{\partial q_{k,n}}{\partial M_{k,n}} = 
\left(-\frac{1}{6}e^{-G(M_{k,n})} - \frac{2}{3}e^{-\frac{4}{3}G(M_{k,n})}\right) 
\frac{\partial G(M_{k,n})}{\partial M_{k,n}}.
\end{equation}
\end{small}

The derivative of \( G(M_{k,n}) \) with respect to \( M_{k,n} \) is calculated as
\begin{small}
\begin{align}
&\frac{\partial G(M_{k,n})}{\partial M_{k,n}} \notag \\
=& \frac{h_k^2 E_k}{N_0 \ln(2) M_{k,n}} 
\left(\sin^2\frac{\pi}{M_{k,n}} - \frac{\pi \log_2(M_{k,n}) \sin\frac{2\pi}{M_{k,n}}}{M_{k,n}}\right).
\end{align}
\end{small}

The latency \( T(M_{k,n}) \) is defined as
\begin{equation}
T(M_{k,n}) = \frac{\phi Z}{B_n \log_2(M_{k,n})}.
\end{equation}

Thus, its derivative is
\begin{equation}
\frac{\partial T}{\partial M_{k,n}} = 
-\frac{\phi Z}{B_n M_{k,n} (\log_2(M_{k,n}))^2 \ln(2)}.
\end{equation}

\subsection*{A.5 Substituting and Simplifying}
Substituting the derivative expressions into the objective derivative, we obtain
\begin{small}
\begin{align}
 &\frac{\lambda\phi Z}{B_n M_{k,n} (\log_2(M_{k,n}))^2 \ln(2)} \notag \\
 =& \frac{D_k^2 \|g_k^t\|^2 \xi}{2\ell D_{all}^2} 
\left[3(\sigma - 1)q_{k,n}^2 + 2q_{k,n}\right]\cdot \notag \\
&\left(-\frac{1}{6}e^{-G(M_{k,n})} - \frac{2}{3}e^{-\frac{4}{3}G(M_{k,n})}\right) 
\frac{\partial G(M_{k,n})}{\partial M_{k,n}}.
\end{align}
\end{small}

This is the final implicit equation for the optimization of modulation order. Then we solve it using the Newton-Raphson method. The algorithm for solving the relaxed modulation order problem is as follows

% \text{Algorithm for Solving the Implicit Equation:}
% \begin{itemize}
% \item Step 1: Initialize modulation order \(M_{k,n}\) with a feasible value.
% \item Step 2: Iteratively update \(M_{k,n}\) using the Newton-Raphson method:
%   \[
%   M^{(t+1)} = M^{(t)} - \frac{f(M^{(t)})}{f'(M^{(t)})}
%   \]
%   where \(f(M)\) is the objective function.
% \item Step 3: Repeat until the change in \(M_{k,n}\) is below a predefined threshold \(\epsilon\), i.e., \(|M^{(t+1)} - M^{(t)}| < \epsilon\).
% \item Step 4: Map the continuous solution to the closest discrete modulation order from the set \(\{2, 4, 8, 16, 64, \dots\}\).
% \end{itemize}

\begin{itemize}
\item Step 1: Initialize $M_{k,n}^{(0)}$ within $[M_{\min},M_{\max}]$ and set the convergence threshold $\epsilon$.
\item Step 2: Compute $f(M_{k,n}^{(i)})=\partial obj(M_{k,n}^{(i)})/\partial M_{k,n}^{(i)}$.
\item Step 3: Update the modulation order by
\[
M_{k,n}^{(i+1)}
=
\Pi_{[M_{\min},M_{\max}]}
\left(
M_{k,n}^{(i)}
-
\frac{f(M_{k,n}^{(i)})}{f'(M_{k,n}^{(i)})}
\right).
\]
\item Step 4: Recalculate $\sigma_{k,n}$ and repeat Steps 2--3 until 
$|M_{k,n}^{(i+1)}-M_{k,n}^{(i)}|<\epsilon$.
\item Step 5: Project the continuous solution onto the feasible discrete modulation set and select the modulation order that maximizes the original objective while satisfying the SER constraint.
\end{itemize}

\section*{Appendix B: Detailed Derivation of KKT Conditions for sub-channel Allocation}
% \section{Detailed Derivation of KKT Conditions for Sub-Channel Allocation}

In this appendix, we derive the KKT conditions for the relaxed sub-channel allocation problem. Given the modulation orders, the relaxed resource allocation problem is formulated as
\begin{small}
\begin{align}
\max_{\mathbf{a},\tau}\quad
& \sum_{k=1}^{K}\sum_{n=1}^{N}a_{k,n}\alpha_{k,n}
-\lambda \tau \\
\mathrm{s.t.}\quad
& T_k^C+\frac{\phi Z}
{\sum_{n=1}^{N}a_{k,n}B_n\log_2(M_{k,n})}
-\tau \le 0,\quad \forall k, \notag\\
& \sum_{k=1}^{K}a_{k,n}-1\le 0,\quad \forall n, \notag\\
& 1-\sum_{n=1}^{N}a_{k,n} \le 0, \quad \forall k, \notag\\
& -a_{k,n}\le 0,\quad \forall k,n, \notag\\
& a_{k,n}-1\le 0,\quad \forall k,n. \notag
\end{align}
\end{small}

For brevity, we define $c_{k,n}=B_n\log_2(M_{k,n}), R_k=\sum_{n=1}^{N}a_{k,n}c_{k,n}$. Then, the latency constraint of client $k$ can be written as $T_k^C+\frac{\phi Z}{R_k}-\tau \le 0$.

To derive the KKT conditions, we convert the maximization problem into an equivalent minimization problem, which is
\begin{align}
\min_{\mathbf{a},\tau}
-\sum_{k=1}^{K}\sum_{n=1}^{N}a_{k,n}\alpha_{k,n}
+\lambda \tau.
\end{align}

Let $\rho_k\ge 0$ be the Lagrange multiplier associated with the latency constraint of client $k$, $\mu_n\ge 0$ be the multiplier associated with the sub-channel exclusiveness constraint, $\underline{\nu}_{k,n}\ge 0$ be the multiplier associated with $-a_{k,n}\le 0$, and $\overline{\nu}_{k,n}\ge 0$ be the multiplier associated with $a_{k,n}-1\le 0$. The Lagrangian function is given by

\begin{small}
\begin{align}
\mathcal{L}
=&
-\sum_{k=1}^{K}\sum_{n=1}^{N}a_{k,n}\alpha_{k,n}
+\lambda \tau \notag \\
&+\sum_{k=1}^{K}\rho_k \left(T_k^C+\frac{\phi Z}{R_k}-\tau
\right) \notag \\
&+ \sum_{n=1}^{N}\mu_n \left( \sum_{k=1}^{K}a_{k,n}-1 \right)+\sum_{k=1}^{K}\omega_k\left(1-\sum_{n=1}^{N}a_{k,n}\right) \notag \\
&-\sum_{k=1}^{K}\sum_{n=1}^{N}\underline{\nu}_{k,n}a_{k,n}
+
\sum_{k=1}^{K}\sum_{n=1}^{N}\overline{\nu}_{k,n}(a_{k,n}-1).
\end{align}
\end{small}

The KKT conditions are given as follows.

\subsection*{B.1 Primal Feasibility}

The primal variables should satisfy
\begin{small}
    \begin{align}
T_k^C+\frac{\phi Z}{R_k}-\tau \le 0,\quad \forall k,\\
\sum_{k=1}^{K}a_{k,n}\le 1,\quad \forall n,\\
\sum_{n=1}^{N}a_{k,n}\ge 1,\quad \forall k,\\
0\le a_{k,n}\le 1,\quad \forall k,n.
\end{align}
\end{small}

\subsection*{B.2 Dual Feasibility}

The Lagrange multipliers should satisfy
\[
\rho_k\ge 0,\quad \mu_n\ge 0,\quad \omega_k \ge 0,\quad
\underline{\nu}_{k,n}\ge 0,\quad
\overline{\nu}_{k,n}\ge 0.
\]

\subsection*{B.3 Complementary Slackness}

The complementary slackness conditions are
\begin{small}
\begin{align}
\rho_k
\left(
T_k^C+\frac{\phi Z}{R_k}-\tau
\right)=0,\quad \forall k,\\
\mu_n\left(\sum_{k=1}^{K}a_{k,n}-1\right)=0,\quad \forall n,\\
\omega_k\left(1-\sum_{n=1}^{N}a_{k,n}\right)=0,\quad \forall k,\\
\underline{\nu}_{k,n}a_{k,n}=0,\quad \forall k,n,\\
\overline{\nu}_{k,n}(a_{k,n}-1)=0,\quad \forall k,n.
\end{align}
\end{small}

\subsection*{B.4 Stationarity}
% Taking the derivative of the Lagrangian with respect to $\tau$, we have
% \begin{align}
% \frac{\partial \mathcal{L}}{\partial \tau}
% =
% \lambda-\sum_{k=1}^{K}\rho_k=0, \sum_{k=1}^{K}\rho_k=\lambda.
% \end{align}

% Next, taking the derivative of the Lagrangian with respect to $a_{k,n}$ gives
% \begin{align}
% \frac{\partial \mathcal{L}}{\partial a_{k,n}}
% =
% -\alpha_{k,n}
% -
% \rho_k
% \frac{\phi Z c_{k,n}}{R_k^2}
% +
% \mu_n
% -\underline{\nu}_{k,n}
% +\overline{\nu}_{k,n}
% =0.
% \end{align}

% Therefore, the stationarity condition can be written as
% \begin{align}
% \alpha_{k,n}
% +
% \rho_k
% \frac{\phi Z c_{k,n}}{R_k^2}
% =
% \mu_n-\underline{\nu}_{k,n}+\overline{\nu}_{k,n}.
% \end{align}

% For an interior solution satisfying $0<a_{k,n}<1$, we have
% \begin{align}
% \underline{\nu}_{k,n}=0,\quad \overline{\nu}_{k,n}=0.
% \end{align}
% Thus, the following relation holds:
% \begin{align}
% \alpha_{k,n}
% +
% \rho_k
% \frac{\phi Z B_n\log_2(M_{k,n})}{R_k^2}
% =
% \mu_n.
% \end{align}

% This condition indicates that the relaxed allocation balances the convergence-related utility $\alpha_{k,n}$ and the latency reduction gain introduced by assigning sub-channel $n$ to client $k$. Specifically, a sub-channel is more likely to be allocated to a client when it provides a larger convergence contribution or a larger reduction in the dominant latency term.

Taking the derivative of the Lagrangian with respect to $\tau$, we have
\begin{align}
\frac{\partial \mathcal{L}}{\partial \tau}=\lambda-\sum_{k=1}^{K}\rho_k=0,\quad \sum_{k=1}^{K}\rho_k=\lambda.
\end{align}

Next, taking the derivative of the Lagrangian with respect to $a_{k,n}$ gives
\begin{align}
\frac{\partial \mathcal{L}}{\partial a_{k,n}}
=
-\alpha_{k,n}
-\rho_k
\frac{\phi Z c_{k,n}}{R_k^2}
+\mu_n
-\omega_k
-\underline{\nu}_{k,n}
+\overline{\nu}_{k,n}
=0,
\end{align}
where $c_{k,n}=B_n\log_2(M_{k,n})$ and
$R_k=\sum_{n=1}^{N}a_{k,n}c_{k,n}$.

Therefore, the stationarity condition can be rewritten as
\begin{align}
\alpha_{k,n}
+
\rho_k
\frac{\phi Z c_{k,n}}{R_k^2}
+
\omega_k
=
\mu_n
-\underline{\nu}_{k,n}
+\overline{\nu}_{k,n}.
\end{align}

For an interior solution satisfying $0<a_{k,n}<1$, we have
\begin{align}
\underline{\nu}_{k,n}=0,\quad \overline{\nu}_{k,n}=0.
\end{align}

Thus, the following relation holds
\begin{align}
\alpha_{k,n}
+
\rho_k
\frac{\phi Z B_n\log_2(M_{k,n})}{R_k^2}
+
\omega_k
=
\mu_n.
\end{align}

This condition indicates that the relaxed allocation balances the convergence-related utility, the latency reduction gain, and the minimum-allocation requirement for each client.

\subsection*{B.5 Binary Recovery}

The KKT conditions provide the optimality characterization of the relaxed sub-channel allocation problem. Since the original problem requires binary sub-channel allocation, the relaxed solution $\tilde a_{k,n}$ is mapped back to a binary solution by
\begin{align}
a_{k,n}^{\star}
=
\begin{cases}
1, & k=\arg\max\limits_{j\in\mathcal{K}}\tilde a_{j,n},\\
0, & \mathrm{otherwise}.
\end{cases}
\end{align}

After the binary recovery, the SER constraint is checked. If the recovered solution violates the SER constraint of any client, the corresponding sub-channel is reassigned to the next feasible client or the modulation order is adjusted to a lower level. This step guarantees that the final solution satisfies the reliability constraint of the original problem.

\bibliographystyle{IEEEtran}
\bibliography{biography.bib}

\vfill

\end{document}